\documentclass[11pt, a4paper]{article}
\usepackage{times, latexsym}
\usepackage{url}
\usepackage[T1]{fontenc}
\usepackage{enumitem}
\usepackage[acceptedWithA]{tacl2021v1}

\usepackage{xspace, mfirstuc, tabulary}

\newif\iftaclinstructions
\taclinstructionsfalse 
\iftaclinstructions
\renewcommand{\confidential}{}
\renewcommand{\anonsubtext}{(No author info supplied here, for consistency with
TACL-submission anonymization requirements)}
\newcommand{\instr}
\fi

\iftaclpubformat 

\else

\fi

\usepackage{amsmath}
\usepackage{multirow}
\usepackage{arydshln}
\usepackage{adjustbox}
\usepackage{booktabs}
\usepackage{graphicx}
\usepackage{inconsolata}
\usepackage{microtype}
\usepackage{amssymb}
\usepackage{xcolor}
\usepackage{pifont}
\usepackage{longtable}
\newcommand{\cmark}{\ding{51}} 
\newcommand{\xmark}{\ding{55}} 

\title{PsyMem: Fine-grained psychological alignment and Explicit Memory Control for
Advanced Role-Playing LLMs}

\author{Xilong Cheng$^{1*}$, Yunxiao Qin$^{1,2,*,\dagger}$, Yuting Tan$^{1,*}$, {Zhengnan Li}$^1$, \\ \textbf{Ye Wang}$^{1,2}$, \textbf{Hongjiang Xiao}$^{1,2,\dagger}$  \textbf{Yuan Zhang}$^{1,2}$\\
	$^1$Communication University of China \\
	$^2$State Key Laboratory of Media Convergence and Communication \\
	\{chengzhengyu330, qinyunxiao, yutingtan, fmlyd, yewang, xiaohj, yzhang\}@cuc.edu.cn\\
}

\date{}

\begin{document}
    \maketitle
    \begin{abstract}
        Existing LLM-based role-playing methods often rely on superficial textual
        descriptions or simplistic metrics, inadequately modeling both intrinsic
        and extrinsic character dimensions. Additionally, they typically
        simulate character memory with implicit model knowledge or basic
        retrieval augment generation without explicit memory alignment, compromising
        memory consistency. The two issues weaken reliability of role-playing
        LLMs in several applications, such as trustworthy social simulation. To
        address these limitations, we propose PsyMem, a novel framework
        integrating fine-grained psychological attributes and explicit memory
        control for role-playing. PsyMem supplements textual descriptions with 26
        psychological indicators to detailed model character. Additionally, PsyMem
        implements memory alignment training, explicitly trains the model to
        align character's response with memory, thereby enabling dynamic memory-controlled
        responding during inference. By training Qwen2.5-7B-Instruct on our specially
        designed dataset (including 5,414 characters and 38,962 dialogues
        extracted from novels), the resulting model, termed as PsyMem-Qwen,
        outperforms baseline models in role-playing, achieving the best performance
        in human-likeness and character fidelity.
    \end{abstract}
    
    \renewcommand{\thefootnote}{\fnsymbol{footnote}}
    \footnotetext[1]{ Equal contribution.}
    \footnotetext[2]{ Corresponding author.}

    \section{Introduction}
    Recent advancements in large language models (LLMs) \citep{achiam2023gpt,anthropic2024claude3,anil2023gemini}
    have unlocked new possibilities for implementing role-playing systems through
    language-based
    behavior simulation \citep{shanahan2023role}. Role-playing systems demonstrate
    societal value across applications like interactive gaming \citep{wang2023voyager}
    and AI counseling \citep{zhang2024cpsycoun}, with particularly potential in enabling high-reliable social simulation. Such simulation allows modeling of human interactions and societal
    dynamics \citep{park2023generative}, offering insights into evidence-based policy design and
    conflict mediation \citep{zhou2024sotopia}. To achieve trustworthy social simulation,
    LLM-based role-playing must preserve persistent traits, memories, and
    behavioral logic to ensure agents reliably mirror specified attributes in
    evolving social contexts.

\begin{figure}[t]
        \includegraphics[width=\columnwidth]{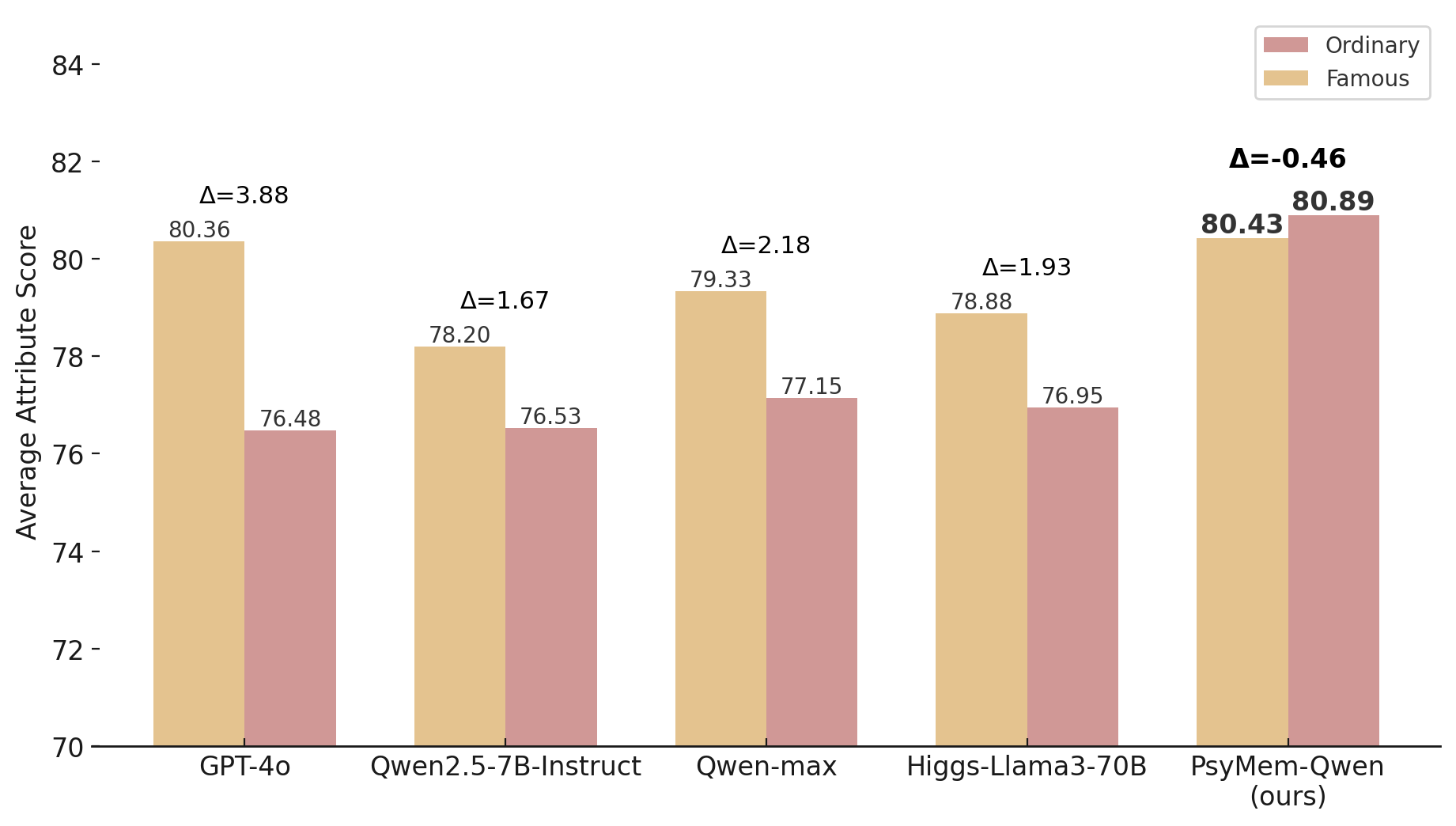}
        \vspace{-30pt}
        \caption{The performance comparison is conducted on two subsets: “Ordinary,”
        consisting of 20 randomly selected characters from our test set, and
        “Famous,” consisting of 20 well-known characters. The average attribute scores
        are calculated as the mean of the quantized scores across the 20 characters
        in each subset.}
        \label{fig:fig2}
    \end{figure}

    Recent works in social or group behavior simulation \citep{kosinski2024evaluating,park2023generative,zhou2024sotopia}
    often rely on general LLMs for role-playing, using context-based learning
    \citep{mann2020language} and instruction follow-up \citep{ouyang2022training}.
    However, these approaches \citep{yu2022xdai} commonly struggle to precisely control
    character attributes due to the unsatisfactory role-playing ability of general
    LLMs, especially for ordinary characters, as Fig.\ref{fig:fig2} shows. Some
    recent works \citep{shao2023character,zhou2023characterglm,tao2023rolecraft}
    proposed the LLMs that are dedicated to role-playing to improve character attribute
    consistency. However, there are still the following two issues: 1) Oversimplified
    characterization and 2) Weak memory control.

    \textbf{The oversimplified characterization issue}. From both
    neuroscientific and psychological perspectives, human attributes can be described
    through two broad dimensions: intrinsic attributes and extrinsic attributes
    \citep{snyder1983influence,kahneman2011thinking}. Intrinsic attributes, such
    as personality traits and values \citep{ravlin1987effect,stumpf1991effects},
    play a significant role in decision-making, particularly in long-term
    decisions \citep{borghans2008economics,roberts2008development}. On the other
    hand, extrinsic attributes \citep{kahneman2011thinking}, such as behavior,
    language, and body language, reflect an individual’s intuitive responses in
    immediate situations \citep{bargh1996automaticity}.

    However, existing role-playing LLMs typically rely on basic textual
    descriptions or narrow metrics to define the target role, which fails to capture
    the full complexity and depth of the character. For example, the work \citep{zhou2023characterglm}
    defines character attributes using textual descriptions based on seven
    dimensions (including identity, interests, experiences, and other relevant
    traits). Some recent research \citep{ran-etal-2024-capturing, yu2024beyond, cui2023machine}
    has attempted to use MBTI for personality trait evaluation to improve character
    control. However, MBTI employs a binary approach to assess personality traits,
    which has low test-retest reliability \citep{stein2019evaluating, hunsley2015controversial}
    and cannot precisely quantify attributes \citep{nowack1996myers}, such as how
    introverted a person is.

    \textbf{The weak memory control issue}. Numerous neuroscientific studies
    have shown that\citep{stumpf1991effects,ravlin1987effect,pickering2004toward},
    during decision-making, the brain relies on not only personality traits (such
    as emotional responses and values) but also long-term memory, particularly in
    fast-paced decision-making contexts. However, existing role-playing models \citep{shao2023character,zhou2023characterglm}
    typically rely on the model’s internal knowledge to simulate character
    memory, which works for famous characters but often fails to provide memory for
    less familiar or unseen ones. Retrieval-Augmented Generation (RAG) \citep{ram2023context}
    has been proposed as a promising solution \citep{zhong2024memorybank}, but we
    observe that simply integrating RAG within Role-Play LLMs' inference stage
    does not adequately address memory management in the context of role-play (see
    Table \ref{tab:model_comparison}).

    In this work, we propose a widely recognized modern psychological evaluation
    framework \citep{costa2008revised,schwartz1992universals,thomas2008thomas,zimbardo1975psychology}
    and a graph-structured memory mechanism to construct the PsyMem framework, enabling
    LLMs to achieve exceptional role-playing abilities. Drawing from modern
    psychology and neuroscience, we enhance the description of character attributes
    by leveraging both \textbf{latent psychological attributes} and \textbf{explicit
    behavioral patterns} \citep{snyder1983influence,kahneman2011thinking}, with personality
    and values corresponding to latent psychological attributes, and behavioral
    decision, social interaction, and leadership representing explicit
    behavioral patterns. Overall, we extract real data from
    novels and construct a large scale role-playing dataset, using 26 quantitative
    indicators and a small amount of textual descriptions to describe characters.
    As shown in Table \ref{tab:datasets_summary}, this dataset contains 5,414 established
    characters, 38,962 conversations, and 536,636 utterances, much larger than
    existing role-playing datasets.

    For character memory, inspired by the hippocampus and cognitive mapping mechanisms
    \citep{eichenbaum2015hippocampus,hartley2014space,tacikowski2024human,tavares2015map},
    we transform characters, events, and relationships into a knowledge graph.
    Unlike previous role playing work \citep{edge2024local} directly using implicit
    model knowledge or integrating role memory in inference, \textbf{we
    incorporate memory information into the training process, forcing the model to
    learn how to respond given retrieved memory}.
    This approach effectively enhances the model’s memory consistency, ensuring that
    role-playing model generates responses based on not only character attributes
    but also memory.

    In general, our five main contributions are as follows:
    \begin{itemize}[leftmargin=9pt]
        \item We propose \textbf{PsyMem}, a novel LLM-based role-playing framework
            that establishes precise control over character behaviors through the
            synergistic integration of fine-grained psychological profiling (e.g.,
            personality, value) and dynamic memory schemata.
            \vspace{-5pt}

        \item Beyond textual descriptions, we present a comprehensive attribute set that bridges intrinsic (e.g., values, emotions) and extrinsic
            (e.g., behaviors, social interactions) character dimensions, operationalized
            via \textbf{26 measurable indicators}. This attribute set captures both
            the internal motivations and external behavioral manifestations of characters.
            \vspace{-5pt}

        \item We introduce a memory alignment training paradigm that establishes
            explicit cognitive anchoring between role-playing responses and character
            memory. This methodology enhances role-playing through a dual-control
            mechanism, enabling \textbf{precise regulation of generative outputs
            via both attributes (e.g., personality dimensions) and episodic
            memory constraints}.
            \vspace{-5pt}

        \item We introduce the use of role-playing style general supervised fine-tuning
            data (transforming general SFT data into role-playing style),
            demonstrating that incorporating such data effectively enhances the
            performance of Role-Playing LLMs.
            \vspace{-5pt}

        \item Extensive experiments demonstrate that the proposed PsyMem framework
            significantly improves the role-playing ability of LLMs. For
            instance, by training Qwen2.5-7B-Instruct on our dataset (introduced
            in Section~\ref{sec: Dataset Architecture}) specially designed for
            PsyMem, the resulting model, termed as PsyMem-Qwen, outperforms baseline
            models in role-playing, achieving the best performance in human-likeness
            and character fidelity.
    \end{itemize}

    \section{Related Work}

    \subsection{LLMs-based Role-Playing}
    In existing role-playing works, the main approaches are Nonparametric
    Prompting and Parametric Training \citep{chen2024persona}. In Nonparametric
    Prompting, \citet{yu2022xdai} use carefully designed character descriptions
    to control general LLMs for role-playing without fine-tuning. However, they
    face significant challenges in accurately reflecting the intrinsic
    relationships between character profiles and dialogue content.

    In Parametric Training, \citet{chen2023large} focus on mimicking characters
    from the Harry Potter series, while \citet{wang2023rolellm} introduced the
    first role-playing dataset with 100 characters generated using GPT-3.5
    prompts. Subsequent works \citep{li2023chatharuhi, zhou2023characterglm, shao2023character}
    have expanded and enriched these datasets by improving the number of
    characters and the diversity of dialogue scenarios, all leveraging various GPT
    models. Meanwhile, \citet{lu-etal-2024-large} adopted a self-alignment
    approach to dataset construction, moving away from the previous method of cheaply
    imitating the role-playing capabilities of GPT models. \citet{zhou2023characterglm}
    leveraged GPT-4 to extract character dialogues from a diverse collection of Chinese
    novels and scripts. More recently, \citet{yu2024beyond} focused on
    addressing the bias between predefined characters and specific scenario
    dialogues, and constructed a dataset of 311 characters.

    Our work belongs to parametric training. We argue that a character's
    behavior is jointly influenced by their attributes, memory, and the scenario.
    Drawing from modern psychology, we segment character attributes using
    contemporary psychological frameworks and construct a dataset containing 38,962
    conversations (536,636 utterances) from real novels, leveraging both character
    attributes and memory to enhance role-playing ability.

    \subsection{Memory Mechanism in Role-Playing.}
    Existing LLMs-based Role-Playing commonly relies on the internal knowledge within
    the model weights to implicitly mimic character memory \citep{zhou2023characterglm,lu-etal-2024-large,li2023chatharuhi}.
    For example, \citet{shao2023character} and others \cite{lu-etal-2024-large, yu2024beyond}
    push the boundaries of knowledge to protect character memory, while \citet{lu-etal-2024-large}
    and others use a self-alignment approach to distinguish the different memory
    backgrounds of various characters. Relying on the internal knowledge within
    the model is effective for famous characters but fails for unfamous and
    unseen characters. Some works apply Retrieval-Augmented Generation (RAG) \citep{ram2023context}
    to dynamically retrieve external contexts for explicitly mimicking character
    memory, such as \citet{zhong2024memorybank} who adopts memory update
    mechanisms based on Ebbinghaus’s forgetting curve theory. However, as shown
    in Table \ref{tab:model_comparison}, existing role-playing models tend to perform
    better when simulating famous characters, while their performance
    significantly drops for ordinary characters that rely on external memory.

    Unlike previous works, we integrate explicit memory control into the training
    framework of role-playing models. This approach encourages the model to
    strictly rely on actual retrieved memories during dialogue generation in
    inference.

    \subsection{Role-playing Evaluation.}
    The current evaluation process consists of dialogue generation and dialogue
    assessment. Due to its efficiency and completeness, automated dialogue
    generation is widely adopted. Evaluation methods are generally categorized into
    three types: Metric-based Evaluation\citep{li2023chatharuhi,wang2023rolellm},
    Human Evaluation\citep{zhou2023characterglm}, and “LLMs as Judges”\citep{shao2023character,lu-etal-2024-large}.
    Metric-based evaluation primarily measures the model’s alignment with standard
    responses, such as original text from novels or manually annotated content. While
    human evaluation is more precise, its high cost and lack of reproducibility
    limit its scalability. In contrast, “LLMs as Judges”, with its efficiency,
    low cost, and strong scalability, is increasingly becoming the preferred
    choice.
    The experiments in Appendix (see Table
    \ref{tab:Cosine similarity between human and GPT-4o scores}) demonstrate the
    effectiveness of the “LLMs as Judges” paradigm for role-playing evaluation.

    \section{Dataset Architecture}
    \label{sec: Dataset Architecture}

    An increasing number of role-playing datasets \citep{ran-etal-2024-capturing,cui2023machine}
    have begun incorporating modern psychological indicators into their characterization
    frameworks. However, previous studies typically introduced only a limited selection
    of psychological attributes as supplementary descriptors, without
    comprehensively integrating psychological systems into character portrayals
    \citep{yu2024beyond,wang2024incharacter}. Moreover, existing datasets predominantly
    focus on character attributes, overlooking the significant influence memory
    has on characters’ behaviors, thought processes, and speech styles\citep{stumpf1991effects,ravlin1987effect,pickering2004toward}.
    This limitation becomes particularly evident when LLMs can hardly access the background knowledge of the role-playing characters during the pretraining stage (see Figure \ref{fig:fig2}).

    To address these shortcomings, we design a novel dataset with the
    architecture formulated as
    \vspace{-5pt}
    \begin{equation}
        \resizebox{0.8\columnwidth}{!}{$\text{D}_{\text{RP}}= \{(R_{i}, P_{i}, M_{i}
        , C_{i}, x_{i}, y_{i}) | i = 1, 2, ..., N\}$}, \label{eq:dataset_RP}
        \vspace{-6pt}
    \end{equation}
    where $N$ is the number of data sample.
    Each sample is a real dialogue turn collected from novels. $R_{i}$ is the
    role for response; 
    $P_{i}$ is the profile of $R_{i}$, including basic information and 26 quantitative psychological dimensions; 
    $C_{i}$ consists of the dialogue turns preceding the current turn, as well as the scene information. 
    $x_{i}$ is the query to $R_{i}$ and $y_{i}$ is the corresponding response. 
    $M_{i}$ denotes the memory retrieved for role $R_{i}$, conditioned on the query $x_{i}$ and the context $C_{i}$, which comprises the preceding dialogue turns and the scene.

\begin{figure}[t]
        \centering
        \includegraphics[width=0.65\columnwidth]{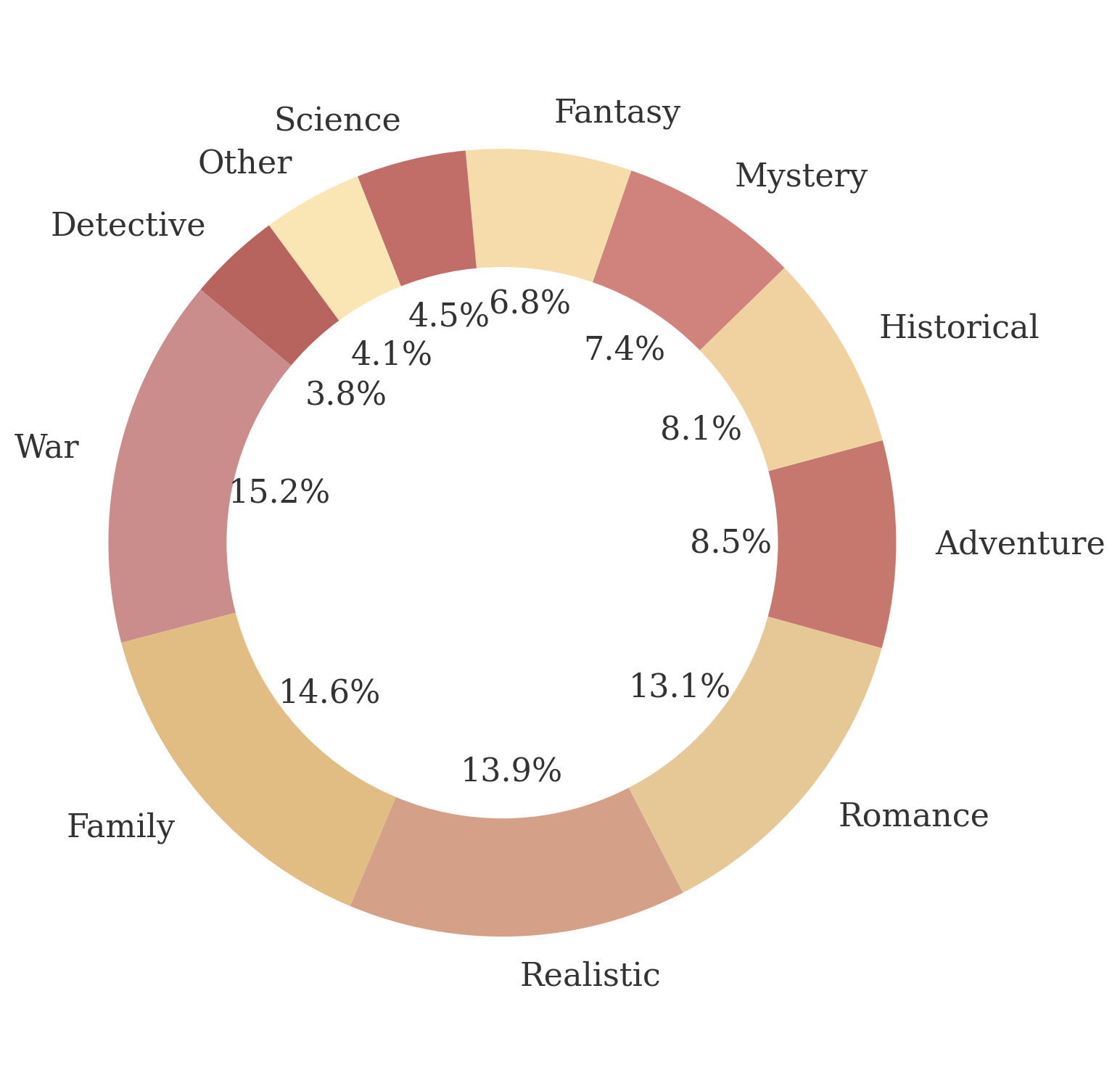}
        \vspace{-25pt}
        \caption{The genre distribution in the dataset.}
        \label{fig:figure5}
        \vspace{-3pt}
    \end{figure}

    In this work, the dataset $\text{D}_{\text{RP}}$ contains 38,962 samples, extracted
    from 539 novels characterized by a wide range of genres, rich content, and narrative
    perspectives of third person.(The genre distribution is presented in Figure
    \ref{fig:figure5}.)

    \begin{table*}
        [t]
        \centering
        \footnotesize
        \setlength{\tabcolsep}{3pt} 
        \begin{tabular}{p{3cm}ccccc cccc}
            \hline
            \multirow{2}{*}{\parbox[t]{3cm}{\raggedright\textbf{ Dataset}}} & \multicolumn{2}{c}{\textbf{Character}} & \multicolumn{3}{c}{\textbf{Conversation}} & \multicolumn{4}{c}{\textbf{Profile Style}} \\
            \cmidrule(lr){2-3} \cmidrule(lr){4-6} \cmidrule(lr){7-10}       & Num.                                   & SA                                        & Conv.                                     & Avg. Turns & Auth.  & PsyFrm & QI     & Text Desc. & Expl. Mem. \\
            \hline
            HPD                                                             & 113                                    & \xmark                                    & 1,191                                     & 13.2       & \cmark & \xmark & \xmark & \cmark     & \xmark     \\
            CharacterGLM                                                    & 250                                    & \xmark                                    & 1,034                                     & 15.8       & \xmark & \xmark & \xmark & \cmark     & \xmark     \\
            RoleLLM                                                         & 100                                    & \xmark                                    & 140,726                                   & 2          & \xmark & \xmark & \xmark & \cmark     & \xmark     \\
            CharacterLLM                                                    & 9                                      & \xmark                                    & 14,300                                    & 13.2       & \xmark & \xmark & \xmark & \cmark     & \xmark     \\
            ChatHaruhi                                                      & 32                                     & \xmark                                    & 54,726                                    & >2         & \xmark & \xmark & \xmark & \cmark     & \xmark     \\
            DITTO                                                           & 4002                                   & \xmark                                    & 7,186                                     & 5.1        & \xmark & \xmark & \xmark & \cmark     & \xmark     \\
            Beyond Dialogue                                                 & 311                                    & \cmark                                    & 3,552                                     & 6.5        & \cmark & \xmark & \cmark & \cmark     & \xmark     \\
            CoSER                                                           & 17,966                                 & \xmark                                    & 29,798                                    & 13.2       & \cmark & \xmark & \xmark & \cmark     & \xmark     \\
            MMRole                                                          & 85                                     & \xmark                                    & 14,346                                    & 4.2        & \xmark & \xmark & \xmark & \cmark     & \xmark     \\
            CharacterBench                                                  & 3956                                   & \xmark                                    & 13,162                                    & 11.3       & \xmark & \xmark & \xmark & \cmark     & \xmark     \\
            CharacterEval                                                   & 77                                     & \xmark                                    & 1,785                                     & 9.3        & \cmark & \xmark & \xmark & \cmark     & \xmark     \\
            \hline
            \textbf{Ours}                                                   & 5,414                                  & \cmark                                    & 38,962                                    & 13.8       & \cmark & \cmark & \cmark & \cmark     & \cmark     \\
            \hline
        \end{tabular}
        \vspace{-8pt}
        \caption{ Dataset Statistics. Comparison between our dataset and
        existing open-source role-play datasets. Columns are categorized into:
        Character (Num.: number of character roles; SA: scene alignment), Conversation
        (Conv.: number of dialogues; Avg. Turns: average dialogue length; Auth.:
        whether dialogues are fully sourced from real, non-generated conversations),
        and Profile Style (PsyFrm: adherence to a psychological framework; QI: presence
        of quantitative indicators; Text Desc.: availability of textual
        character descriptions; Expl. Mem.: support for explicit memory
        retrieval during interaction). }
        \vspace{-5pt}
        \label{tab:datasets_summary}
    \end{table*}

    \begin{figure*}[t]
        \includegraphics[width=0.96\linewidth]{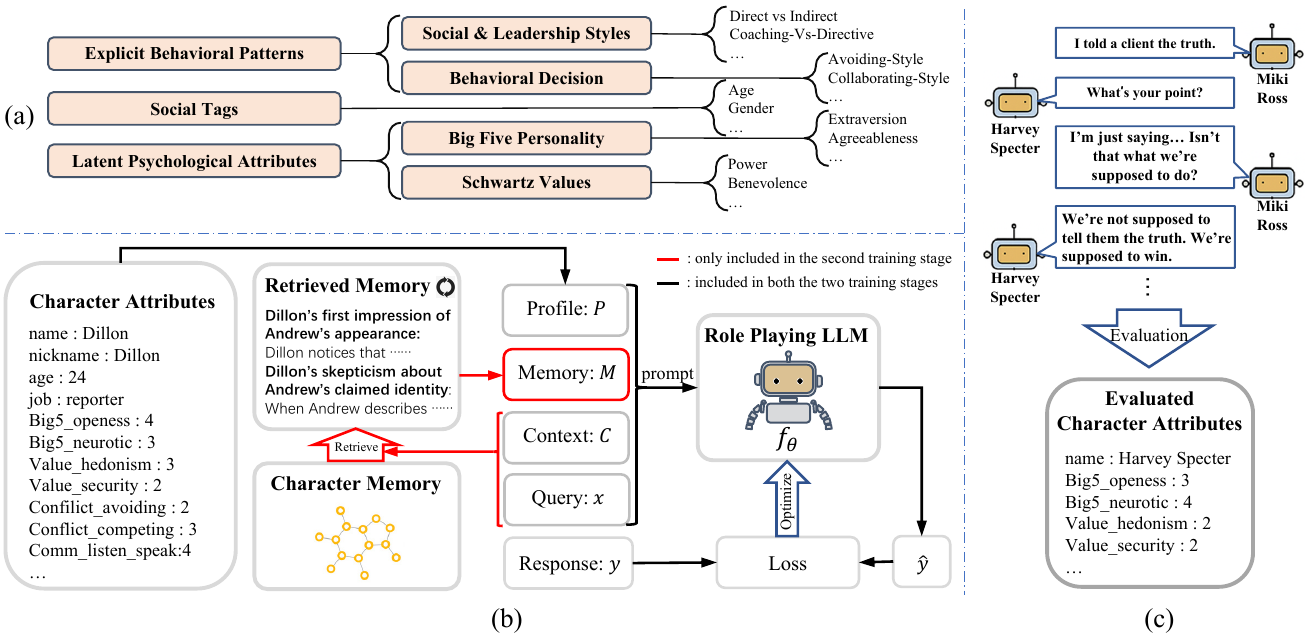}
        \vspace{-13pt}
        \caption{(a): The fine-grained character attributes designed in this work.
        (b): The two stages training of role-playing LLM. First, the model learns
        basic role-playing without character memory. In the second stage, we dynamically
        retrieves memory relevant to the current query and dialogue context from
        graph-structured character memory. We then integrate character profile, retrieved
        memories, dialogue history, and current input to enhance role-playing precision,
        training the model to align responses with both fine-grained character
        profile and contextual memory. (c):We assess the role-playing LLM by first
        generating a multi-turn dialogue (up to 15 turns) between two designated
        roles, followed by three rounds of scoring with GPT-4o based on a
        quantitative rubric, and report the mean score. }
        \vspace{-5pt}
        \label{Figure: sft-and-evaluation}
    \end{figure*}

    \subsection{Quantifiable Character Attributes}
    \label{subsec: Quantifiable Character Attributes}

    We systematically construct quantifiable character attribute dimensions aligned
    with contemporary psychological frameworks by categorizing character traits into
    \textbf{latent psychological attributes} and \textbf{explicit behavioral
    patterns} \citep{kahneman2011thinking,snyder1983influence}, as illustrated in
    Figure \ref{Figure: sft-and-evaluation}a. These two dimensions distinctly influence
    individual behavior, communication styles, decision-making processes, and
    interpersonal interactions. Besides, a small amount of text describing the
    social attributes of the characters is also included as a supplement.

    \noindent
    \textbf{Latent psychological attributes} reflect internal, often
    subconscious aspects of character, guiding motivations, preferences, and emotional
    responses. To describe these latent psychological attributes comprehensively,
    we utilize the following psychological frameworks:

    \textbf{1) Personality}: Evaluated using the Big Five Personality Model \citep{costa2008revised},
    encompassing stable emotional, cognitive, and behavioral tendencies. Traits
    such as openness, conscientiousness, extraversion, agreeableness, and
    neuroticism directly influence communication styles, social interactions,
    and general life choices.

    \textbf{2) Values}: Assessed through Schwartz's Theory of Basic Values
    \citep{schwartz1992universals}, which classifies core beliefs into ten
    universally recognized types. These values deeply affect decision-making,
    priorities, ethical judgments, and interpersonal alignment, enriching the psychological
    depth and motivational dimensions of individuals.

    \vspace{5pt}
    \noindent
    \textbf{Explicit behavioral patterns} manifest in observable actions, interactions,
    and communication styles. We capture these explicit patterns using:

    \textbf{1) Social \& Leadership Styles}: Measured through six dimensions based
    on Zimbardo's psychological principles \citep{zimbardo1975psychology}, significantly
    influencing observable interpersonal behaviors, communication effectiveness,
    leadership approaches, and social dynamics in groups.

    \textbf{2) Behavioral Decision}: Analyzed using the Thomas-Kilmann Conflict
    Mode Instrument (TKI) \citep{thomas2008thomas}, emphasizing decision-making
    processes when characters face conflicts or challenges. This explicitly captures
    how emotional, social, and heuristic influences shape specific decisions,
    providing precise and actionable insights into behavioral responses.

    \subsection{Character Memory Integration}

    Neuroscientific research underscores the significant role memory plays in shaping
    human behavior, communication, and decision-making processes \citep{stumpf1991effects,ravlin1987effect,pickering2004toward}.
    Memories influence not only how individuals perceive and interpret events but
    also guide their interactions and responses within social contexts,
    highlighting the importance of accurately capturing and utilizing memory in
    character portrayal.

    Therefore, we introduce character memory data $M$ to the role playing dataset to enable the memory simulation capability of role playing model.
    Specifically, we first build a knowledge graph for each novel via GraphRAG \citep{edge2024local}, and then retrieve relevant memory context for the current dialogue in local mode (see Appendix~\ref{sub: Memory Module} for details).
    Memory alignment is evaluated on two key dimensions: \textit{correctness} and \textit{rationality}. 
    Correctness assesses whether the model accurately utilizes stored character information in response to queries, while rationality examines logical coherence and contextual appropriateness. 
    Additionally, simple summaries or vague memory references are identified as non-rational uses of memory.
    To further enhance robustness, irrelevant memory items are deliberately introduced as noise, promoting stable, realistic, and consistent character portrayals reflective of human memory dynamics.

    In addition, we define character-specific knowledge boundaries to ensure consistency—for
    example, Quasimodo from The Hunchback of Notre-Dame would not be aware of
    torch. Based on the absence of relevant retrievable memories, we used GPT-4o
    to generate 800 refusal-style QA pairs for 100 characters, grounded in real-world
    constraints and narrative context. These were integrated into novel-based
    dialogues to maintain coherence and character fidelity.

    \subsection{Synthetic Role-playing Data}
    In addition to $\text{D}_{\text{RP}}$, we propose a synthetic role-playing data
    construction approach that transforms general supervised fine-tuning (SFT) data
    into role-playing style dialogues, resulting in the dataset
    $\text{D}_{\text{RP}}^{\text{synth}}$. The construction pipeline consists of
    three steps: 1) We randomly sample multi-turn dialogues from the Pure-Dove (PD)
    dataset \citep{daniele2023amplify-instruct} and sequentially group every 100
    consecutive dialogue turns to form speaker-specific clusters. Each cluster containing
    approximately 100 dialogue turns and is then assigned to represent a distinct
    synthetic speaker. 2) For each synthetic speaker, we use GPT-4o to analyze
    the 100 dialogue turns and annotate the speaker's psychological profile
    according to our 26 quantitative dimensions. 3) Based on the dialogue
    context, we employ GPT-4o to generate appropriate scene-specific information
    to enhance contextual coherence. The resulting dataset
    $\text{D}_{\text{RP}}^{\text{synth}}$ is formulated as:
    \vspace{-3pt}
    \begin{equation}
        \resizebox{0.8\columnwidth}{!}{$\text{D}_{\text{RP}}^{\text{synth}}= \{(R
        _{i}, P_{i}, C_{i}, x_{i}, y_{i}) | i = 1, 2, ..., N\}$}, \label{eq:dataset_RB}
        \vspace{-3pt}
    \end{equation}
    where $R_{i}$ denotes the role responsible for the response, and $P_{i}$
    is the profile of $R_{i}$ defined by basic information and 26 quantitative psychological dimensions.
    $C_{i}$ represents the preceding turns and scene information.
    $x_{i}$ and $y_{i}$ are the current query, and the corresponding response drawn from the Pure-Dove
    (PD) datasets, respectively.

    By incorporating role-playing style general SFT data $\text{D}_{\text{RP}}^{\text{synth}}$,
    the proposed PsyMem framework enables the model to develop stronger role-playing
    capabilities, as Figure \ref{fig:Figure4} shows.

    \section{Role Playing Training and Evaluation}
    \vspace{-2pt}
    \subsection{Training}
    \vspace{-2pt}
    In this work, we transfer general LLMs to role-playing models via supervised
    fine-tuning. Role-playing presents a unique challenge for language models: maintaining
    both general capabilities and character-specific attributes simultaneously.
    This challenge exemplifies the fundamental tension in transfer learning
    between preserving general knowledge and specializing for domain-specific
    tasks. 
    Traditional fine-tuning approaches \citep{kotha2024understanding} often encounter a core multi-task conflict during optimization. In the context of role-playing tasks, this conflict arises from the attempt to optimize two fundamentally distinct objectives—namely, character profile alignment and memory adherence—within a unified framework.
    Inspired by the Dual-stage Mixed Fine-tuning (DMT) \citep{dong2023abilities} method, we introduce a two-stage training strategy for role-playing, effectively balancing the capabilities of character profile alignment and memory adherence.

    Specifically, stage 1 establishes core role-playing capabilities by training the model on character profile-based dialogues, enabling the model to learn fundamental character attribute alignment. 
    Stage 2 further trains the model on memory-augmented role-playing dataset $\text{D}_{\text{RP}}$ and role-playing style general fine-tuning dataset $\text{D}_{\text{RP}}^\text{synth}$ to improve role-playing precision, as illustrated in Figure \ref{Figure: sft-and-evaluation}(b). 
    As demonstrated in Section \ref{sec:Ablation Study}, our two-stage training strategy significantly outperforms vanilla single-stage training for role-playing performance, validating the necessity of this progressive specialization strategy.

    \paragraph{Stage 1: Foundational Role-Playing Capacity Development}
    In this stage,we train LLMs to learn foundational role-playing abilities
    based on quantifiable character attributes and dialogue contexts without
    memory augmentation. We fine-tune the model using a subset $\text{D}_{\text{RP1}}$
    sampled from the overall role-playing dataset $\text{D}_{\text{RP}}$ to establish
    its fundamental role-playing abilities. The objective function is formally
    defined as:
    \begin{equation}
        \label{eq:stage1_objective}\resizebox{0.85\columnwidth}{!}{$\theta^{*}_{1}
        = \arg\min_{\theta}\mathbb{E}_{(\cdot) \sim \text{D}_\text{RP1}}\left[ \mathcal{L}
        (f_{\theta}(R,P,C,x), y) \right]$},
    \end{equation}
    where $(\cdot)$ in $\text{D}_{\text{RP1}}$ denotes the tuple $(R,P,C,x,y)$ and
    $\text{D}_{\text{RP1}}$ excludes the retrieved memory component $M$ from the
    original dataset $\text{D}_{\text{RP}}$.

    \textbf{Stage 2: Memory-Augmented Role-Playing Specialization.} In this stage, we enhance the model's role-playing capabilities through two training components: 1) memory-augmented role-playing data $\text{D}_{\text{RP}}$, which incorporates long-term memory contexts retrieved via graph-based methods to enable explicit memory alignment, and (2) synthetic role-playing style data {$\text{D}_{\text{RP}}^{\text{synth}}$}, which improves general role-playing abilities. 
    This stage employs a weighted combination of two corresponding loss terms to simultaneously enhance memory-controlled response generation while preserving the model's foundational capabilities:
    \vspace{-6pt}
    \begin{equation}
        \label{eq:stage2_objective_total}\resizebox{0.8\columnwidth}{!}{{$\theta^{*}_{2} = \arg\min_{\theta}\left\{ \alpha \mathcal{L}_{\mathrm{RP}}(\theta) + \mathcal{L}_{\mathrm{RP}}^{\mathrm{synth}}(\theta) \right\}$}}
        \vspace{-5pt}
    \end{equation}
    where $\alpha = 20$.

    The two loss terms are individually defined as follows:
    \vspace{-6pt}
    \begin{equation}
        \label{eq:stage2_rp_loss}\resizebox{0.8\columnwidth}{!}{$\mathcal{L}_{\mathrm{RP}}
        (\theta) = \mathbb{E}_{(\cdot) \sim \text{D}_\text{RP}}\left[\mathcal{L}\left
        (f_{\theta}(R,P,M,C,x), y\right)\right]$},
        \vspace{-6pt}
    \end{equation}
    \vspace{-6pt}
    \begin{equation}
        \label{eq:stage2_rb_loss}\resizebox{0.8\columnwidth}{!}{$\mathcal{L}_{\mathrm{RP}}
        ^{\mathrm{synth}}(\theta) = \mathbb{E}_{(\cdot) \sim \text{D}_\text{RP}^\text{synth}}
        \left[\mathcal{L}\left(f_{\theta}(R,P,C,x), y\right)\right]$}
        \vspace{-3pt}
    \end{equation}
    where $(\cdot)$ in $\text{D}_{\text{RP}}^{\text{synth}}$ represents
    $(R,P,C,x,y)$ and $(\cdot)$ in $\text{D}_{\text{RP}}$ represents the tuple $(R,P,M,C,x,y)$. 
    $M$ denotes the retrieved memory response for role $R$ regarding the query $x$ and the context $C$ (which includes the preceding dialogue turns and scene information), obtained via graph-based retrieval methods and subsequently filtered by LLMs.

    \begin{table*}
        [t]
        \centering
        \scriptsize
        \setlength{\tabcolsep}{2.0pt}
        \begin{tabular}{@{}lccccccccc@{}}
            \toprule \multirow{2}{*}{\textbf{Model}}                                           & \multicolumn{6}{c}{\textbf{Character Fidelity}} & \multicolumn{3}{c}{\textbf{Character-independent}} \\
            \cmidrule(lr){2-7} \cmidrule(l){8-10}                                              & \textbf{Per.}                                   & \textbf{Val.}                                     & \textbf{SL}$^{*}$                          & \textbf{BD}$^{*}$                          & \textbf{Mem.}                     & \textbf{Avg.}                              & \textbf{H-like}$^{*}$                       & \textbf{Cons.}                    & \textbf{Coh.}                    \\
            \midrule \noalign{\vskip -3pt} \multicolumn{1}{@{}l}{\tiny\textit{General LLMs}}   &                                                 &                                                   &                                            &                                            &                                   & \multicolumn{1}{c}{}                       &                                             &                                   &                                  \\
            \noalign{\vskip -3.5pt} \midrule GPT-4o                                            & 79.08                                           & 80.11                                             & 77.43                                      & 73.29                                      & 94.40                             & 80.86                                      & 66.60                                       & 89.28                             & \textbf{99.20}                   \\
            GPT-3.5-Turbo                                                                      & 79.08                                           & 79.42                                             & 74.15                                      & 72.11                                      & 84.60                             & 77.87                                      & 34.40                                       & 88.60                             & 96.20                            \\
            Claude-3.5-sonnet                                                                  & 80.34                                           & 80.29                                             & 78.51                                      & 74.57                                      & 91.80                             & 81.10                                      & 75.00                                       & 89.50                             & \textbf{99.20}                   \\
            Qwen-Max                                                                           & 78.67                                           & 80.42                                             & 76.54                                      & 72.37                                      & 90.60                             & 79.72                                      & 57.60                                       & 89.48                             & 97.60                            \\
            Yi-Large                                                                           & 80.20                                           & 80.09                                             & 78.27                                      & 76.82                                      & 92.00                             & 81.48                                      & 83.80                                       & 88.18                             & 98.40                            \\
            Deepseek-R1                                                                        & 80.98                                           & 81.38                                             & 77.88                                      & 77.12                                      & \textbf{95.00}                    & 82.47                                      & 77.20                                       & 87.57                             & 96.20                            \\
            Deepseek-V3                                                                        & 79.82                                           & 80.33                                             & 79.16                                      & 75.08                                      & 93.40                             & 81.56                                      & 76.40                                       & 89.04                             & 99.00                            \\
            LLaMA3.1-8B-instruct                                                               & 79.17                                           & 80.14                                             & 76.76                                      & 76.01                                      & 83.20                             & 79.06                                      & 64.20                                       & 88.74                             & 99.00                            \\
            Qwen2.5-7B-Instruct                                                                & 77.44                                           & 80.02                                             & 76.15                                      & 76.51                                      & 85.60                             & 79.14                                      & 64.40                                       & 89.58                             & 97.40                            \\
            \midrule \noalign{\vskip -3pt} \multicolumn{10}{@{}l}{\tiny\textit{Role-play LLMs}} \\
            \noalign{\vskip -3.5pt} \midrule CharacterGLM-6B                                   & {80.66}                                         & 80.67                                             & 79.35                                      & 71.20                                      & 40.20                             & 70.42                                      & 52.80                                       & 89.61                             & 73.00                            \\
            Baichuan-NPC-Turbo                                                                 & 76.53                                           & 79.05                                             & 74.29                                      & 72.27                                      & 87.20                             & 77.87                                      & 61.60                                       & \textbf{90.36}                    & 96.40                            \\
            Hunyuan-Role                                                                       & 79.31                                           & 79.71                                             & 78.24                                      & 78.38                                      & 82.00                             & 79.53                                      & 87.40                                       & 89.16                             & 91.20                            \\
            Higgs-LLaMA3-70B                                                                   & 80.02                                           & 80.61                                             & 75.77                                      & 74.08                                      & 93.20                             & 80.74                                      & 60.40                                       & 88.40                             & 99.60                            \\
            CoSER-70B                                                                          & \textbf{81.21}                                  & 81.90                                             & 78.83                                      & 76.35                                      & 93.00                             & 82.28                                      & 81.80                                       & 89.44                             & 97.20                            \\
            \midrule PsyMem-LLaMA (\textbf{ours})                                              & 80.47$^{\textcolor{blue}{+1.30}}$               & \textbf{81.93}$^{\textcolor{blue}{+1.79}}$        & 78.57$^{\textcolor{blue}{+1.81}}$          & 74.47$^{\textcolor{red}{-1.54}}$           & 90.20$^{\textcolor{blue}{+7.00}}$ & 81.13$^{\textcolor{blue}{+2.07}}$          & 85.20$^{\textcolor{blue}{+21.00}}$          & 89.93$^{\textcolor{blue}{+1.19}}$ & 95.60$^{\textcolor{red}{-3.40}}$ \\
            PsyMem-Qwen (\textbf{ours})                                                        & 80.40$^{\textcolor{blue}{+2.96}}$               & 81.74$^{\textcolor{blue}{+1.72}}$                 & \textbf{80.80}$^{\textcolor{blue}{+4.65}}$ & \textbf{78.48}$^{\textcolor{blue}{+1.97}}$ & 91.80$^{\textcolor{blue}{+6.20}}$ & \textbf{82.64}$^{\textcolor{blue}{+3.50}}$ & \textbf{87.60}$^{\textcolor{blue}{+23.20}}$ & 89.64$^{\textcolor{blue}{+0.06}}$ & 96.20$^{\textcolor{red}{-1.20}}$ \\
            \bottomrule
        \end{tabular}
        \vspace{-12pt}
        \caption{Model evaluation across multiple criteria. \textbf{Bold}
        indicates best performance. Superscripts show change from base model.
        Abbreviations: Per. = Personality, Val. = Values, SL$^{*}$ = Social \&
        Leadership, BD$^{*}$ = Behavioral Decision, Mem. = Memory, H-like$^{*}$
        = Human-likeness, Cons. = Consistency, Coh. = Coherence }
        \label{tab:model_comparison}
        \vspace{-5pt}
    \end{table*}

    \subsection{Evaluation}
    \label{subsec: Evaluation}

    We evaluate role-playing capabilities of LLMs by adopting the widely used
    “LLMs as Judges” approach\footnote{We validate the reliability of our LLM-based evaluation framework by comparing human-evaluation and GPT-4o judgments in Appendix (see Table \ref{tab:Cosine similarity between human and GPT-4o scores})} \citep{shao2023character,lu-etal-2024-large}, as illustrated in Figure \ref{Figure: sft-and-evaluation}(c). 
    The evaluation process is structured into two main categories: \textbf{Character-independent Capabilities} and \textbf{Character Fidelity}. 

    \paragraph{Character-independent Capabilities.}
    Following previous research \citep{chen2024persona,yu2024beyond} , we
    measure Character-independent Capabilities using three metrics: Human-likeness,
    Consistency and Coherence. In the Human-likeness task, we use GPT-4o with
    few-shot examples to determine whether a dialogue was generated by a human, ultimately
    computing the accuracy score. For consistency, we evaluate the coherence between
    different parts of the conversation to assess whether the role-playing remains
    consistent throughout long dialogues. For coherence, we assess how logically
    connected and contextually appropriate the responses are across the entire dialogue,
    ensuring the conversation flows naturally.

    \paragraph{Character Fidelity.}
    Character Fidelity assesses how accurately an LLM portrays specific role-playing
    characters. As described in Section~\ref{subsec: Quantifiable Character Attributes},
    we employ modern psychological quantification frameworks defined in our dataset
    design, covering five primary dimensions: Personality (Big Five Personality
    Model), Values (Schwartz’s Theory of Basic Values), Social and Leadership Styles
    (Zimbardo’s psychological principles), Behavioral Decision (Thomas-Kilmann Conflict
    Mode Instrument, TKI), and Memory. The former four dimensions are quantified
    into numeric scales ranging from 1 to 5 using standardized psychological evaluation
    methods.
    The Memory dimension, distinct from the psychological attributes, is
    evaluated using accuracy scores from true/false assessments.

    \paragraph{Evaluation Process.}
    The evaluation process commences with models generating 15-turn dialogues
    per scenario (see Section \ref{sec:experimental setup} for scenario detail),
    embodying the specified character profile.
    \textit{For character-independent capabilities evaluation}, we utilize these
    generated dialogues: \textbf{1)} Human-likeness and Coherence are assessed
    by GPT-4o (true/false based on rules), yielding a positive assessment rate (e.g.,
    the percentage deemed human-like or coherent).
    \textbf{2)} Consistency is measured by the score difference in the four psychological
    dimensions between dialogue halves, yielding a stability score (e.g., where a
    smaller difference yields a higher consistency score, scaled from 0 to 1).
    \textit{For character fidelity evaluation}, GPT-4o serves as the judge, assessing
    each 15-turn dialogue. Performance is scored across five dimensions: \textbf{1)}
    The first four dimensions (Personality, Values, etc.,) are evaluated by GPT-4o
    on dialogues, using 1-NED (introduced in the following paragraph) for performance;
    \textbf{2)} For the fifth dimension, memory alignment, we use GPT-4o to judge
    response correctness (true/false) against memory context and reference answer,
    yielding an accuracy score (i.e., the proportion of correctly aligned responses).

    \paragraph{Metrics.}
    We quantify role-playing fidelity for Personality, Values, Social \& Leadership,
    and Behavioral Decision using the 1-NED metric. Derived from Normalized Euclidean
    Distance (NED), a higher 1-NED score indicates greater similarity to the
    target profile.
    The 1-NED score is computed as:
    \vspace{-8pt}
    \begin{equation}
        \label{eq:normalized_euclidean}\resizebox{0.8\columnwidth}{!}{$\mathbf{1-NED}
        = 1 - \frac{1}{\Delta X}\sqrt{\frac{1}{n}\sum_{i=1}^{n}\left(X_{i} - X_{\mathrm{ref}}\right)^{2}}$}
        ,
        \vspace{-3pt}
    \end{equation}
    where $\Delta \! X =\!\! X_{\max}\!\!-\! X_{\min}$,
    with $X_{\max}$ and $X_{\min}$ being the maximum (5) and minimum (1) values of
    the evaluation scale, respectively. A score of 0, used to denote uncertainty,
    is excluded from this range. $n$ is the number of subdimensions, $X_{i}$ represents
    the measurement for the $i$-th subdimension, and $X_{\mathrm{ref}}$ is the
    corresponding value contained in role profile $P_{i}$ (see Eq.(\ref{eq:dataset_RP})
    and Eq.(\ref{eq:dataset_RB})), with possible values from $\{1, 2, 3, 4, 5\}$.
    This formulation is robust, as it normalizes the mean squared deviation across
    different scales, yielding a more interpretable similarity metric.

    \section{Experiments}
    \vspace{-3pt}
    In this section, we explain the dataset, baselines, and implementation, and
    discuss the main results and ablation experiments.

    \subsection{Experimental Setup}
    \vspace{-3pt}
    \label{sec:experimental setup}
    \paragraph{Evaluation Dataset.}
    To mitigate potential knowledge bias introduced during the pretraining phase, we construct an evaluation dataset $\text{D}_{\text{RP}}^{\text{eval}}$ for our main experiments by selecting characters and scenes from novels published after June 2024. 
    This design choice aims to ensure that models have had minimal or no exposure to the selected texts during pretraining, thereby reducing reliance on memorized character-specific knowledge.

    We first assess 100 contemporary novels across diverse genres, then select 25 high-quality novels based on character complexity, dialogue richness, expression quality, and narrative perspective. 
    From each selected novel, we identify the two most psychologically complex and representative characters, yielding 50 characters with varied psychological profiles. 
    For each character, we extract ten dialogue-rich scenes directly from the original novels, resulting in 500 evaluation scenarios (50 characters × 10 scenes),  which constitutes our evaluation set $\text{D}_{\text{RP}}^{\text{eval}}$. 
    The character attributes are derived from a detailed analysis of the novel using GPT-4o.
    This comprehensive evaluation framework enables robust assessment of role-playing capabilities across diverse character types and narrative contexts. 
    For the memory evaluation, each scenario includes not only the character’s profile and scene information, but also three key components: a meticulously selected character memory excerpt from the memory text, a memory-dependent question, and its authentic answer serving as the reference answer, allowing us to assess all character fidelity dimensions.

    \paragraph{Baselines.}
    Baselines in our evaluation include both open-source and proprietary
    advanced chatbots, each representing the state-of-the-art within their respective
    frameworks. For general-purpose models, we compared the proposed PsyMem with
    several prominent baselines, including GPT-4o \citep{hurst2024gpt}, GPT-3.5-turbo 
    \citep{openai2023gpt35turbo},Claude-3.5-sonnet \citep{anthropic2023claude}, Qwen-Max 
    \citep{qwen25}, Yi-Large \citep{young2024yi}, Deepseek-R1 \citep{guo2025deepseek},
    DeepSeek-V3 \citep{liu2024deepseek}, LLaMA3.1-8B-instruct \citep{grattafiori2024LLaMA}
    and Qwen2.5-7B-Instruct \citep{yang2024qwen2}. These models are considered
    some of the best in the field, offering a broad range of capabilities across
    various domains. In the category of Role-Playing-focused baselines, we made comparisons
    with Hunyuan-Role\citep{tencent2023}, Baichuan-NPC-Turbo\citep{baichuan2023},
    and CharacterGLM-6B \citep{zhou2023characterglm},Higgs-LLaMA3-70B\citep{boson2025higgs},CoSER-70B\citep{wang2025cosercoordinatingllmbasedpersona},
    which are specifically designed to handle character-driven interactions and story-based
    tasks.

    \paragraph{Implementations.}
    With the proposed PsyMem framework, we fine-tune LLaMA3.1-8B-Instruct
    and Qwen2.5-7B-Instruct for three epochs, and denote the resulting models as
    PsyMem-LLaMA and PsyMem-Qwen, respectively. Additional training settings and
    implementation details are provided in the Appendix~\ref{sub:Hyperparameters}. Evaluations are conducted according to Section~\ref{subsec: Evaluation}.

    \subsection{Main Results}
    We present the main results in Table \ref{tab:model_comparison}.
    Within the General Baselines, we observe that proprietary models commonly outperform
    open-source models. Claude-3.5-sonnet and Yi-Large, as outstanding proprietary
    models, exhibit impressive performance in Character-independent Capabilities.
    Additionally, Baichuan-NPC-Turbo, as a role-play expertise baseline model,
    surpasses general baselines in self-consistency, which is particularly important
    for role-playing agents. Furthermore, it is worth mentioning that DeepSeek-R1
    and CoSER-70B excel in Character Fidelity.

    Empirical results under the PsyMem framework reveal that, relative to their
    original counterparts, PsyMem-LLaMA and PsyMem-Qwen achieve character fidelity
    improvements of 2.07\% and 3.50\%, respectively. In addition, PsyMem-Qwen
    demonstrates strong performance in the Social \& Leadership and Behavioral
    Decision dimensions, achieving scores of 80.80\% and 78.48\%, respectively. PsyMem-LLaMA,
    on the other hand, obtains a score of 81.93\% in the Value dimension, surpassing
    all other evaluated models in this category. \textbf{Moreover, PsyMem-Qwen,
    despite having only 7B parameters, surpassed all measured baselines in Character
    Fidelity.} Additionally, it is worth noting that both models exhibited
    remarkable improvements in Human-likeness. Overall, the PsyMem framework has
    demonstrated strong effectiveness in our benchmark tests, enabling precise character
    control through quantitative character attribute metrics and memory mechanisms.

    \begin{table}
        \centering
        \resizebox{\columnwidth}{!}{%
        \begin{tabular}{l|cccccc}
            \hline
            \multirow{2}{*}{\textbf{Model}}       & \multicolumn{6}{c}{\textbf{Character Fidelity}} \\
            \cline{2-7}                           & \textbf{Per.}                                  & \textbf{Val.}  & \textbf{SL$^{*}$} & \textbf{BD$^{*}$} & \textbf{Mem.}  & {\textbf{Avg.}} \\
            \hline
            \textbf{LLaMA3.1-8B-Instruct}         &                                                &                &                   &                   &                &                 \\
            + PsyMem (w/o Mem.)                   & 80.11                                          & 81.90          & \textbf{79.23}    & \textbf{76.36}    & 44.20          & 72.36           \\
            + PsyMem (w/o DMT.)                   & 79.97                                          & 81.34          & 77.96             & 74.55             & 89.40          & 80.64           \\
            + PsyMem                              & \textbf{80.47}                                 & \textbf{81.93} & 78.57             & 74.47             & \textbf{90.20} & \textbf{81.13}  \\
            \midrule \textbf{Qwen2.5-7B-Instruct} &                                                &                &                   &                   &                &                 \\
            + PsyMem (w/o Mem.)                   & \textbf{81.02}                                 & 81.33          & \textbf{82.10}    & \textbf{79.02}    & 53.00          & 75.29           \\
            + PsyMem (w/o DMT.)                   & 79.59                                          & 81.55          & 80.37             & 77.96             & 90.40          & 81.97           \\
            + PsyMem                              & 80.40                                          & \textbf{81.74} & 80.80             & 78.48             & \textbf{91.80} & \textbf{82.64}  \\
            \hline
        \end{tabular}%
        }
        \vspace{-10pt}
        \caption{Ablation studies on DMT. and memory integration
        within the PsyMem framework.
        }
        \label{tab:ablation_study}
    \end{table}

    \subsection{Ablation Study}
    \label{sec:Ablation Study}

    \begin{table}[t]
        \centering
        \resizebox{\columnwidth}{!}{%
        \begin{tabular}{l|c|c|c|c}
            \hline
            \multirow{2}{*}{\textbf{Model}}                     & \textbf{Character} & \textbf{Human}     & \textbf{Consis-} & \textbf{Coher-} \\
                                                                & \textbf{Fidelity}  & \textbf{-likeness} & \textbf{tency}   & \textbf{ence}   \\
            \hline
            \textbf{LLaMA3.1-8B-Instruct}                       & 79.06              & 64.20              & 88.74            & \textbf{99.00}  \\
            ~~~+ PsyMem (PD data)                               & 80.51              & 70.80              & 89.49            & 82.20           \\
            ~~~+ PsyMem ($\text{D}_{\text{RP}}^{\text{synth}}$) & \textbf{81.13}     & \textbf{85.20}     & \textbf{89.93}   & 95.60           \\
            \midrule \textbf{Qwen2.5-7B-Instruct}               & 79.14              & 64.40              & 89.58            & \textbf{97.40}  \\
            ~~~+ PsyMem (PD data)                               & 81.63              & 86.60              & 89.45            & 93.40           \\
            ~~~+ PsyMem ($\text{D}_{\text{RP}}^{\text{synth}}$) & \textbf{82.64}     & \textbf{87.60}     & \textbf{89.64}   & 96.20           \\
            \hline
        \end{tabular}%
        }
        \vspace{-10pt}
        \caption{Ablation study on Synthetic Role-playing style ($\text{D}_{\text{RP}}
        ^{\text{synth}}$) Pure Dove (PD) dataset in PsyMem.}
        \label{tab:ablation_rb_pd}
        \vspace{-5pt}
    \end{table}

    \begin{figure}[t]
        \includegraphics[width=\columnwidth]{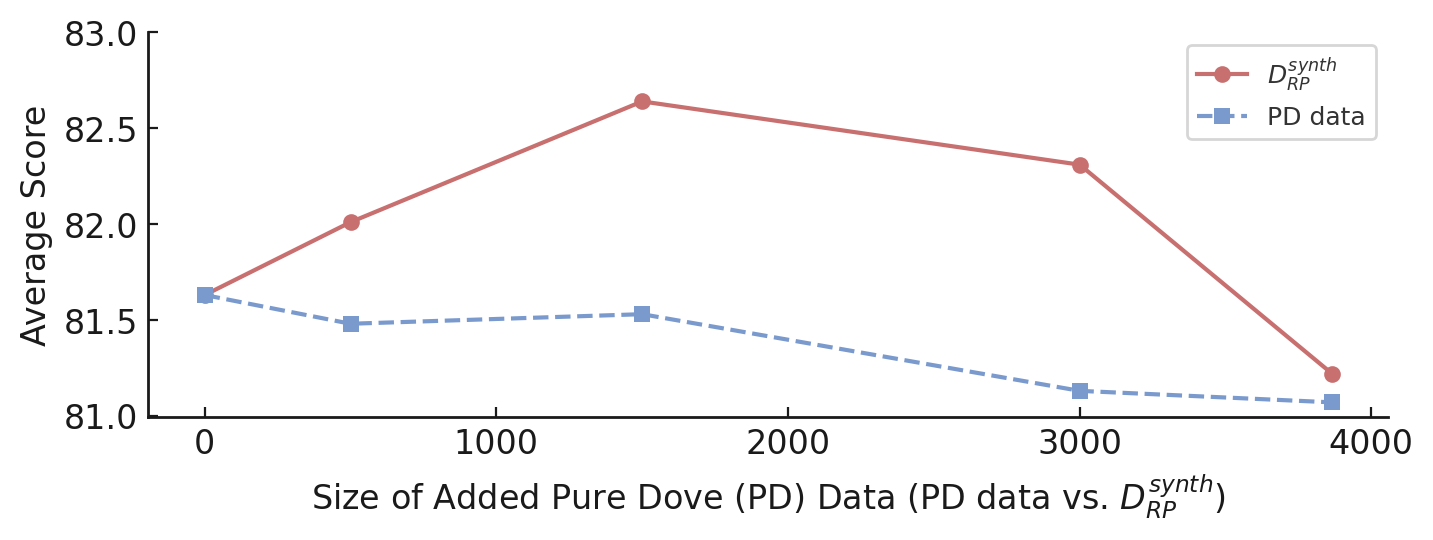}
        \vspace{-25pt}
        \caption{Comparing Original and Synthetic Role-Playing data ($\text{D}_{\text{RP}}
        ^{\text{synth}}$). PD: Performance by Dataset Size.}
        \label{fig:Figure4}
    \end{figure}

    \textbf{Memory alignment training.} To evaluate the effectiveness of memory alignment
    training, we compare the performance of PsyMem-LLaMA and PsyMem-Qwen with
    and without memory integration during the second training stage. 
    The results are presented in Table \ref{tab:ablation_study}. 
    \textbf{Our findings indicate that incorporating memory data in the second training stage
    significantly enhances the model’s ability to responding based on retrieved memory, rather than merely summarizing relevant content. } 
    This improves both character attribute consistency and long-term memory retention. 
    However, we also observe that memory alignment training has a slight negative impact on character attribute alignment. 
    We hypothesize that this may be due to the inclusion of a limited number of role-specific dialogue examples in the memory data, which could slightly diminish the model’s ability to adhere strictly to predefined character attributes.

    \vspace{5pt}
    \textbf{Two-stage Training Strategy.}
    To evaluate the effectiveness of the DMT strategy, which serves as the core of
    our two-stage training strategy approach, we conducted ablation studies
    comparing the full PsyMem framework (with DMT) against a single-stage fine-tuning
    baseline, in which all datasets and loss functions are jointly optimized. As
    shown in Table~\ref{tab:ablation_study}, DMT consistently improves performance
    across both backbone models (Qwen and LLaMA). For PsyMem-Qwen, DMT leads to an
    overall improvement of 0.67\% (82.64\% vs. 81.97\%), while for PsyMem-LLaMA,
    the improvement is 0.49\% points (81.13\% vs. 80.64\%), \textbf{demonstrating
    the effectiveness of the two-stage approach in enhancing character fidelity.}

    \vspace{5pt}
    \textbf{{Synthetic Role-playing Data.}} We investigate the impact of
    Synthetic Role-playing style SFT Data $\text{D}_{\text{RP}}^{\text{synth}}$
    on model performance through ablation studies presented in Table
    \ref{tab:ablation_rb_pd} and Figure \ref{fig:Figure4}.

    Table \ref{tab:ablation_rb_pd} compares model performance using
    $\text{D}_{\text{RP}}^{\text{synth}}$ versus original general SFT data. The results
    demonstrate that incorporating \textbf{$\text{D}_{\text{RP}}^{\text{synth}}$
    significantly improves Character Fidelity, Human-likeness, and Consistency
    for both PsyMem-LLaMA and PsyMem-Qwen}, while effectively maintaining coherence
    capabilities.

    Figure \ref{fig:Figure4} further examines the effect of data volume,
    comparing General SFT Data with $\text{D}_{\text{RP}}^{\text{synth}}$ on
    role-playing performance. The results validate the superiority of our synthetic
    approach: while General SFT Data consistently degrades character-specific
    abilities, Synthetic Role-playing style data achieves optimal performance at
    1,500 samples before declining at larger volumes. We attribute this
    performance drop at 3,000 samples to dataset distribution imbalance, where excessive
    synthetic content may dilute role-specific attribute diversity.

    \begin{figure}[t]
        \includegraphics[width=\columnwidth]{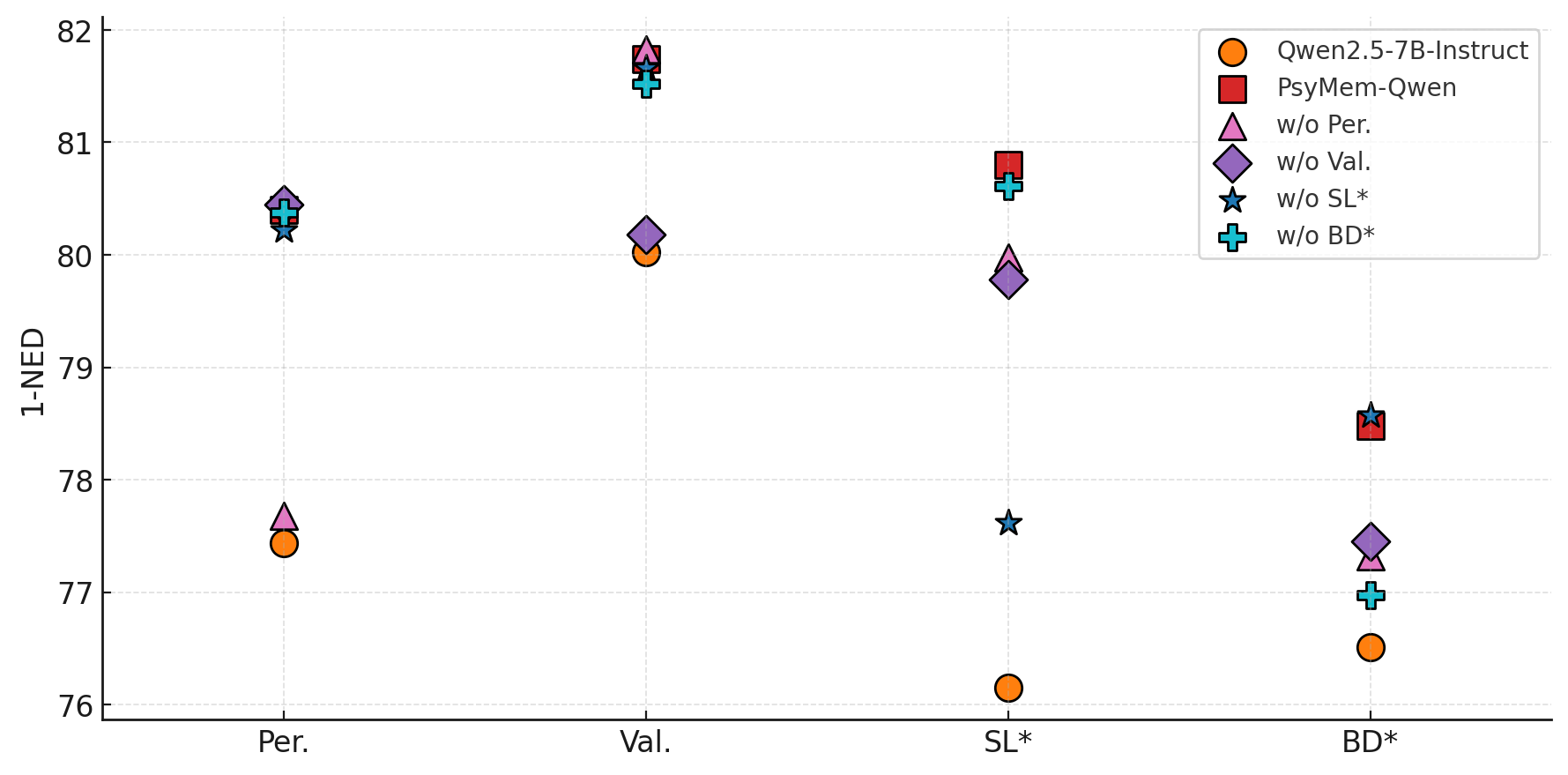}
        \vspace{-25pt}
        \caption{Ablation study of each dimensions. Per.: Personality, Val.:
        Values, SL$^{*}$: Social \& Leadership, BD$^{*}$: Behavioral Decision, Mem.:
        Memory.}
        \label{fig:ablation_chart}
        \vspace{-5pt}
    \end{figure}

    \vspace{5pt}
    \textbf{Character Attribute Dimension.} We assess the
    contribution of each character attribute by conducting an ablation study on
    PsyMem-Qwen, where each of the four supervision signals—Personality, Values,
    Social \& Leadership, and Behavioral Decision—is individually removed (see Figure~\ref{fig:ablation_chart}).
    The results reveal that removing any single dimension markedly drops performance
    on its corresponding evaluation metric. This indicates each dimension is
    critical to the model's fidelity for that trait, underscoring the necessity of
    the proposed comprehensive attribute set. We have placed more ablation experiments on dimensions in Appendix~\ref{sub:Evaluation-stage Input Ablation}.

    \section{Conclusion}
    This paper presented PsyMem, a novel framework for LLM-based role-playing that
    synergizes character memory with fine-grained psychological attributes,
    aligning with contemporary psychological theories. Evaluations conducted on a
    large-scale novel-based dataset demonstrate PsyMem's effectiveness in role-playing.
    The resulting model, PsyMem-Qwen, even with only 7B size parameters, outperforms
    all baselines in character fidelity while also demonstrating strong character-independent
    capabilities.
    Additionally, we highlighted the advantages of integrating Synthetic Role-playing
    style General SFT Data. We believe this work paves the way for future
    research into psychologically-informed role-playing models, fostering more nuanced,
    realistic, and consistent character portrayals and thereby expanding the utility
    of LLMs in applications such as trustworthy social simulation, enhanced
    human-computer interaction, and AI counseling.

    \bibliographystyle{acl_natbib}
    \bibliography{main}

\begin{thebibliography}{72}
\expandafter\ifx\csname natexlab\endcsname\relax\def\natexlab#1{#1}\fi

\bibitem[{Achiam et~al.(2023)Achiam, Adler, Agarwal, Ahmad, Akkaya, Aleman,
  Almeida, Altenschmidt, Altman, Anadkat et~al.}]{achiam2023gpt}
Josh Achiam, Steven Adler, Sandhini Agarwal, Lama Ahmad, Ilge Akkaya,
  Florencia~Leoni Aleman, Diogo Almeida, Janko Altenschmidt, Sam Altman,
  Shyamal Anadkat, et~al. 2023.
\newblock Gpt-4 technical report.
\newblock \emph{arXiv preprint arXiv:2303.08774}.

\bibitem[{AI(2024)}]{boson2025higgs}
Boson AI. 2024.
\newblock \href {https://boson.ai/higgs-opensource/} {Announcing the higgs
  family of llms}.

\bibitem[{Anil et~al.(2023)Anil, Borgeaud, Wu, Alayrac, Yu, Soricut, Schalkwyk,
  Dai, Hauth, Millican et~al.}]{anil2023gemini}
Rohan Anil, Sebastian Borgeaud, Yonghui Wu, Jean-Baptiste Alayrac, Jiahui Yu,
  Radu Soricut, Johan Schalkwyk, Andrew~M Dai, Anja Hauth, Katie Millican,
  et~al. 2023.
\newblock Gemini: A family of highly capable multimodal models.
\newblock \emph{arXiv preprint arXiv:2312.11805}, 1.

\bibitem[{Anthropic(2024{\natexlab{a}})}]{anthropic2024claude3}
Anthropic. 2024{\natexlab{a}}.
\newblock \href
  {https://www-cdn.anthropic.com/de8ba9b01c9ab7cbabf5c33b80b7bbc618857627/Model_Card_Claude_3.pdf}
  {The claude 3 model family: Opus, sonnet, haiku}.
\newblock Accessed: 2025-02-14.

\bibitem[{Anthropic(2024{\natexlab{b}})}]{anthropic2023claude}
Anthropic. 2024{\natexlab{b}}.
\newblock \href {https://www.anthropic.com/news/claude-3-5-sonnet} {Claude 3.5
  sonnet}.

\bibitem[{{Baichuan AI}(2023)}]{baichuan2023}
{Baichuan AI}. 2023.
\newblock \href {https://npc.baichuan-ai.com/index} {Baichuan npc platform}.
\newblock Accessed: 2025-02-16.

\bibitem[{Bargh et~al.(1996)Bargh, Chen, and Burrows}]{bargh1996automaticity}
John~A Bargh, Mark Chen, and Lara Burrows. 1996.
\newblock Automaticity of social behavior: Direct effects of trait construct
  and stereotype activation on action.
\newblock \emph{Journal of personality and social psychology}, 71(2):230.

\bibitem[{{BookLikes}(2024)}]{booklikes-home}
{BookLikes}. 2024.
\newblock \href {https://booklikes.com/} {Booklikes – a community for book
  lovers}.
\newblock Accessed: 2025-06-22.

\bibitem[{Borghans et~al.(2008)Borghans, Duckworth, Heckman, and
  Ter~Weel}]{borghans2008economics}
Lex Borghans, Angela~Lee Duckworth, James~J Heckman, and Bas Ter~Weel. 2008.
\newblock The economics and psychology of personality traits.
\newblock \emph{Journal of human Resources}, 43(4):972--1059.

\bibitem[{Chen et~al.(2024)Chen, Wang, Xu, Yuan, Zhang, Shi, Xie, Li, Yang, Zhu
  et~al.}]{chen2024persona}
Jiangjie Chen, Xintao Wang, Rui Xu, Siyu Yuan, Yikai Zhang, Wei Shi, Jian Xie,
  Shuang Li, Ruihan Yang, Tinghui Zhu, et~al. 2024.
\newblock From persona to personalization: A survey on role-playing language
  agents.
\newblock \emph{arXiv preprint arXiv:2404.18231}.

\bibitem[{Chen et~al.(2023)Chen, Wang, Jiang, Cai, Li, Chen, Wang, and
  Li}]{chen2023large}
Nuo Chen, Yan Wang, Haiyun Jiang, Deng Cai, Yuhan Li, Ziyang Chen, Longyue
  Wang, and Jia Li. 2023.
\newblock Large language models meet harry potter: A dataset for aligning
  dialogue agents with characters.
\newblock In \emph{Findings of the Association for Computational Linguistics:
  EMNLP 2023}, pages 8506--8520.

\bibitem[{Costa and McCrae(2008)}]{costa2008revised}
Paul~T Costa and Robert~R McCrae. 2008.
\newblock The revised neo personality inventory (neo-pi-r).
\newblock \emph{The SAGE handbook of personality theory and assessment},
  2(2):179--198.

\bibitem[{Cui et~al.(2023)Cui, Lv, Wen, Wang, Tang, Tian, and
  Yuan}]{cui2023machine}
Jiaxi Cui, Liuzhenghao Lv, Jing Wen, Rongsheng Wang, Jing Tang, Yonghong Tian,
  and Li~Yuan. 2023.
\newblock Machine mindset: An mbti exploration of large language models.
\newblock \emph{arXiv preprint arXiv:2312.12999}.

\bibitem[{Daniele and Suphavadeeprasit(2023)}]{daniele2023amplify-instruct}
Luigi Daniele and Suphavadeeprasit. 2023.
\newblock \href {https://huggingface.co/datasets/LDJnr/Capybara}
  {Amplify-instruct: Synthetically generated diverse multi-turn conversations
  for efficient llm training.}
\newblock \emph{arXiv preprint arXiv:(coming soon)}.

\bibitem[{{DeepMind}(2025)}]{deepmind2025gemini2.5}
{DeepMind}. 2025.
\newblock Gemini 2.5: Pushing the frontier with advanced reasoning,
  multimodality, long context, and next generation agentic capabilities.
\newblock
  \url{https://storage.googleapis.com/deepmind-media/gemini/gemini_v2_5_report.pdf}.
\newblock Technical report.

\bibitem[{Dong et~al.(2023)Dong, Yuan, Lu, Li, Xue, Liu, Wang, Yuan, Zhou, and
  Zhou}]{dong2023abilities}
Guanting Dong, Hongyi Yuan, Keming Lu, Chengpeng Li, Mingfeng Xue, Dayiheng
  Liu, Wei Wang, Zheng Yuan, Chang Zhou, and Jingren Zhou. 2023.
\newblock How abilities in large language models are affected by supervised
  fine-tuning data composition.
\newblock \emph{arXiv preprint arXiv:2310.05492}.

\bibitem[{Edge et~al.(2024)Edge, Trinh, Cheng, Bradley, Chao, Mody, Truitt, and
  Larson}]{edge2024local}
Darren Edge, Ha~Trinh, Newman Cheng, Joshua Bradley, Alex Chao, Apurva Mody,
  Steven Truitt, and Jonathan Larson. 2024.
\newblock From local to global: A graph rag approach to query-focused
  summarization.
\newblock \emph{arXiv preprint arXiv:2404.16130}.

\bibitem[{Eichenbaum(2015)}]{eichenbaum2015hippocampus}
Howard Eichenbaum. 2015.
\newblock The hippocampus as a cognitive map… of social space.
\newblock \emph{Neuron}, 87(1):9--11.

\bibitem[{Fleiss(1971)}]{fleiss1971measuring}
Joseph~L Fleiss. 1971.
\newblock Measuring nominal scale agreement among many raters.
\newblock \emph{Psychological bulletin}, 76(5):378.

\bibitem[{{Goodreads}(2024)}]{goodreads-fiction}
{Goodreads}. 2024.
\newblock \href {https://www.goodreads.com/genres/fiction} {Goodreads}.
\newblock Accessed: 2025-06-22.

\bibitem[{Grattafiori et~al.(2024)Grattafiori, Dubey, Jauhri, Pandey, Kadian,
  Al-Dahle, Letman, Mathur, Schelten, Vaughan et~al.}]{grattafiori2024LLaMA}
Aaron Grattafiori, Abhimanyu Dubey, Abhinav Jauhri, Abhinav Pandey, Abhishek
  Kadian, Ahmad Al-Dahle, Aiesha Letman, Akhil Mathur, Alan Schelten, Alex
  Vaughan, et~al. 2024.
\newblock The llama 3 herd of models.
\newblock \emph{arXiv preprint arXiv:2407.21783}.

\bibitem[{Guo et~al.(2025)Guo, Yang, Zhang, Song, Zhang, Xu, Zhu, Ma, Wang, Bi
  et~al.}]{guo2025deepseek}
Daya Guo, Dejian Yang, Haowei Zhang, Junxiao Song, Ruoyu Zhang, Runxin Xu,
  Qihao Zhu, Shirong Ma, Peiyi Wang, Xiao Bi, et~al. 2025.
\newblock Deepseek-r1: Incentivizing reasoning capability in llms via
  reinforcement learning.
\newblock \emph{arXiv preprint arXiv:2501.12948}.

\bibitem[{Hartley et~al.(2014)Hartley, Lever, Burgess, and
  O'Keefe}]{hartley2014space}
Tom Hartley, Colin Lever, Neil Burgess, and John O'Keefe. 2014.
\newblock Space in the brain: how the hippocampal formation supports spatial
  cognition.
\newblock \emph{Philosophical Transactions of the Royal Society B: Biological
  Sciences}, 369(1635):20120510.

\bibitem[{Hauke and Kossowski(2011)}]{hauke2011comparison}
Jan Hauke and Tomasz Kossowski. 2011.
\newblock Comparison of values of pearson's and spearman's correlation
  coefficients on the same sets of data.
\newblock \emph{Quaestiones geographicae}, 30(2):87--93.

\bibitem[{Hunsley et~al.(2015)Hunsley, Lee, Wood, and
  Taylor}]{hunsley2015controversial}
John Hunsley, Catherine~M Lee, James~M Wood, and Whitney Taylor. 2015.
\newblock Controversial and questionable assessment techniques.

\bibitem[{Hurst et~al.(2024)Hurst, Lerer, Goucher, Perelman, Ramesh, Clark,
  Ostrow, Welihinda, Hayes, Radford et~al.}]{hurst2024gpt}
Aaron Hurst, Adam Lerer, Adam~P Goucher, Adam Perelman, Aditya Ramesh, Aidan
  Clark, AJ~Ostrow, Akila Welihinda, Alan Hayes, Alec Radford, et~al. 2024.
\newblock Gpt-4o system card.
\newblock \emph{arXiv preprint arXiv:2410.21276}.

\bibitem[{Kahneman(2011)}]{kahneman2011thinking}
Daniel Kahneman. 2011.
\newblock Thinking, fast and slow.
\newblock \emph{Farrar, Straus and Giroux}.

\bibitem[{Kosinski(2024)}]{kosinski2024evaluating}
Michal Kosinski. 2024.
\newblock Evaluating large language models in theory of mind tasks.
\newblock \emph{Proceedings of the National Academy of Sciences},
  121(45):e2405460121.

\bibitem[{Kotha et~al.(2024)Kotha, Springer, and
  Raghunathan}]{kotha2024understanding}
Suhas Kotha, Jacob~Mitchell Springer, and Aditi Raghunathan. 2024.
\newblock \href {https://openreview.net/forum?id=VrHiF2hsrm} {Understanding
  catastrophic forgetting in language models via implicit inference}.
\newblock In \emph{The Twelfth International Conference on Learning
  Representations}.

\bibitem[{Krippendorff(2018)}]{krippendorff2018content}
Klaus Krippendorff. 2018.
\newblock \emph{Content analysis: An introduction to its methodology}.
\newblock Sage publications.

\bibitem[{Li et~al.(2023)Li, Leng, Yan, Shen, Wang, Mi, Fei, Feng, Yan, Wang
  et~al.}]{li2023chatharuhi}
Cheng Li, Ziang Leng, Chenxi Yan, Junyi Shen, Hao Wang, Weishi Mi, Yaying Fei,
  Xiaoyang Feng, Song Yan, HaoSheng Wang, et~al. 2023.
\newblock Chatharuhi: Reviving anime character in reality via large language
  model.
\newblock \emph{arXiv preprint arXiv:2308.09597}.

\bibitem[{Liu et~al.(2024)Liu, Feng, Xue, Wang, Wu, Lu, Zhao, Deng, Zhang, Ruan
  et~al.}]{liu2024deepseek}
Aixin Liu, Bei Feng, Bing Xue, Bingxuan Wang, Bochao Wu, Chengda Lu, Chenggang
  Zhao, Chengqi Deng, Chenyu Zhang, Chong Ruan, et~al. 2024.
\newblock Deepseek-v3 technical report.
\newblock \emph{arXiv preprint arXiv:2412.19437}.

\bibitem[{Lu et~al.(2024)Lu, Yu, Zhou, and Zhou}]{lu-etal-2024-large}
Keming Lu, Bowen Yu, Chang Zhou, and Jingren Zhou. 2024.
\newblock \href {https://doi.org/10.18653/v1/2024.acl-long.423} {Large language
  models are superpositions of all characters: Attaining arbitrary role-play
  via self-alignment}.
\newblock In \emph{Proceedings of the 62nd Annual Meeting of the Association
  for Computational Linguistics (Volume 1: Long Papers)}, pages 7828--7840,
  Bangkok, Thailand. Association for Computational Linguistics.

\bibitem[{Mann et~al.(2020)Mann, Ryder, Subbiah, Kaplan, Dhariwal, Neelakantan,
  Shyam, Sastry, Askell, Agarwal et~al.}]{mann2020language}
Ben Mann, N~Ryder, M~Subbiah, J~Kaplan, P~Dhariwal, A~Neelakantan, P~Shyam,
  G~Sastry, A~Askell, S~Agarwal, et~al. 2020.
\newblock Language models are few-shot learners.
\newblock \emph{arXiv preprint arXiv:2005.14165}, 1.

\bibitem[{Marzi et~al.(2024)Marzi, Balzano, and Marchiori}]{marzi2024k}
Giacomo Marzi, Marco Balzano, and Davide Marchiori. 2024.
\newblock K-alpha calculator--krippendorff's alpha calculator: a user-friendly
  tool for computing krippendorff's alpha inter-rater reliability coefficient.
\newblock \emph{MethodsX}, 12:102545.

\bibitem[{Nowack(1996)}]{nowack1996myers}
K~Nowack. 1996.
\newblock Is the myers briggs type indicator the right tool to use.
\newblock \emph{Performance in Practice}, 6.

\bibitem[{{OpenAI}(2023)}]{openai2023gpt35turbo}
{OpenAI}. 2023.
\newblock \href {https://platform.openai.com/docs/models/gpt-3-5-turbo}
  {Gpt-3.5 turbo documentation}.
\newblock Accessed: 2025-02-16.

\bibitem[{{OpenAI}(2025)}]{openai2025o3o4mini}
{OpenAI}. 2025.
\newblock Introducing openai o3 and o4‑mini.
\newblock \url{https://openai.com/index/introducing-o3-and-o4-mini/}.
\newblock Accessed YYYY-MM-DD.

\bibitem[{Ouyang et~al.(2022)Ouyang, Wu, Jiang, Almeida, Wainwright, Mishkin,
  Zhang, Agarwal, Slama, Ray et~al.}]{ouyang2022training}
Long Ouyang, Jeffrey Wu, Xu~Jiang, Diogo Almeida, Carroll Wainwright, Pamela
  Mishkin, Chong Zhang, Sandhini Agarwal, Katarina Slama, Alex Ray, et~al.
  2022.
\newblock Training language models to follow instructions with human feedback.
\newblock \emph{Advances in neural information processing systems},
  35:27730--27744.

\bibitem[{Park et~al.(2023)Park, O'Brien, Cai, Morris, Liang, and
  Bernstein}]{park2023generative}
Joon~Sung Park, Joseph O'Brien, Carrie~Jun Cai, Meredith~Ringel Morris, Percy
  Liang, and Michael~S Bernstein. 2023.
\newblock Generative agents: Interactive simulacra of human behavior.
\newblock In \emph{Proceedings of the 36th annual acm symposium on user
  interface software and technology}, pages 1--22.

\bibitem[{Pickering and Garrod(2004)}]{pickering2004toward}
Martin~J Pickering and Simon Garrod. 2004.
\newblock Toward a mechanistic psychology of dialogue.
\newblock \emph{Behavioral and brain sciences}, 27(2):169--190.

\bibitem[{Ram et~al.(2023)Ram, Levine, Dalmedigos, Muhlgay, Shashua,
  Leyton-Brown, and Shoham}]{ram2023context}
Ori Ram, Yoav Levine, Itay Dalmedigos, Dor Muhlgay, Amnon Shashua, Kevin
  Leyton-Brown, and Yoav Shoham. 2023.
\newblock In-context retrieval-augmented language models.
\newblock \emph{Transactions of the Association for Computational Linguistics},
  11:1316--1331.

\bibitem[{Ran et~al.(2024)Ran, Wang, Xu, Yuan, Liang, Xiao, and
  Yang}]{ran-etal-2024-capturing}
Yiting Ran, Xintao Wang, Rui Xu, Xinfeng Yuan, Jiaqing Liang, Yanghua Xiao, and
  Deqing Yang. 2024.
\newblock \href {https://doi.org/10.18653/v1/2024.findings-emnlp.853}
  {Capturing minds, not just words: Enhancing role-playing language models with
  personality-indicative data}.
\newblock In \emph{Findings of the Association for Computational Linguistics:
  EMNLP 2024}, pages 14566--14576, Miami, Florida, USA. Association for
  Computational Linguistics.

\bibitem[{Ravlin and Meglino(1987)}]{ravlin1987effect}
Elizabeth~C Ravlin and Bruce~M Meglino. 1987.
\newblock Effect of values on perception and decision making: A study of
  alternative work values measures.
\newblock \emph{Journal of Applied psychology}, 72(4):666.

\bibitem[{Roberts et~al.(2008)Roberts, Wood, and
  Caspi}]{roberts2008development}
Brent~W Roberts, Dustin Wood, and Avshalom Caspi. 2008.
\newblock The development of personality traits in adulthood.

\bibitem[{Schwartz(1992)}]{schwartz1992universals}
Shalom~H Schwartz. 1992.
\newblock Universals in the content and structure of values: Theoretical
  advances and empirical tests in 20 countries.
\newblock \emph{Advances in experimental social psychology/Academic Press}.

\bibitem[{Shanahan et~al.(2023)Shanahan, McDonell, and
  Reynolds}]{shanahan2023role}
Murray Shanahan, Kyle McDonell, and Laria Reynolds. 2023.
\newblock Role play with large language models.
\newblock \emph{Nature}, 623(7987):493--498.

\bibitem[{Shao et~al.(2023)Shao, Li, Dai, and Qiu}]{shao2023character}
Yunfan Shao, Linyang Li, Junqi Dai, and Xipeng Qiu. 2023.
\newblock Character-llm: A trainable agent for role-playing.
\newblock \emph{arXiv preprint arXiv:2310.10158}.

\bibitem[{Snyder(1983)}]{snyder1983influence}
Mark Snyder. 1983.
\newblock The influence of individuals on situations: Implications for
  understanding the links between personality and social behavior.
\newblock \emph{Journal of personality}, 51(3):497--516.

\bibitem[{Stein and Swan(2019)}]{stein2019evaluating}
Randy Stein and Alexander~B Swan. 2019.
\newblock Evaluating the validity of myers-briggs type indicator theory: A
  teaching tool and window into intuitive psychology.
\newblock \emph{Social and Personality Psychology Compass}, 13(2):e12434.

\bibitem[{Stumpf and Dunbar(1991)}]{stumpf1991effects}
Stephen~A Stumpf and Roger~LM Dunbar. 1991.
\newblock The effects of personality type on choices made in strategic decision
  situations.
\newblock \emph{Decision Sciences}, 22(5):1047--1072.

\bibitem[{Tacikowski et~al.(2024)Tacikowski, Kalender, Ciliberti, and
  Fried}]{tacikowski2024human}
Pawel Tacikowski, G{\"u}ldamla Kalender, Davide Ciliberti, and Itzhak Fried.
  2024.
\newblock Human hippocampal and entorhinal neurons encode the temporal
  structure of experience.
\newblock \emph{Nature}, 635(8037):160--167.

\bibitem[{Tao et~al.(2023)Tao, Liang, Shi, Yu, and Xie}]{tao2023rolecraft}
Meiling Tao, Xuechen Liang, Tianyu Shi, Lei Yu, and Yiting Xie. 2023.
\newblock Rolecraft-glm: Advancing personalized role-playing in large language
  models.
\newblock \emph{arXiv preprint arXiv:2401.09432}.

\bibitem[{Tavares et~al.(2015)Tavares, Mendelsohn, Grossman, Williams, Shapiro,
  Trope, and Schiller}]{tavares2015map}
Rita~Morais Tavares, Avi Mendelsohn, Yael Grossman, Christian~Hamilton
  Williams, Matthew Shapiro, Yaacov Trope, and Daniela Schiller. 2015.
\newblock A map for social navigation in the human brain.
\newblock \emph{Neuron}, 87(1):231--243.

\bibitem[{Team(2024)}]{qwen25}
Qwen Team. 2024.
\newblock Qwen2.5 technical report.
\newblock \emph{arXiv preprint arXiv:2412.15115}.

\bibitem[{{Tencent Cloud}(2023)}]{tencent2023}
{Tencent Cloud}. 2023.
\newblock \href {https://cloud.tencent.com/document/product/1729/104753}
  {Document for tencent cloud product 1729}.
\newblock Accessed: 2025-02-16.

\bibitem[{{The StoryGraph}(2024)}]{storygraph-browse}
{The StoryGraph}. 2024.
\newblock \href {https://app.thestorygraph.com/browse} {The storygraph}.
\newblock Accessed: 2025-06-22.

\bibitem[{Thomas(2008)}]{thomas2008thomas}
Kenneth~W Thomas. 2008.
\newblock Thomas-kilmann conflict mode.
\newblock \emph{TKI Profile and Interpretive Report}, 1(11).

\bibitem[{Wang et~al.(2023{\natexlab{a}})Wang, Xie, Jiang, Mandlekar, Xiao,
  Zhu, Fan, and Anandkumar}]{wang2023voyager}
Guanzhi Wang, Yuqi Xie, Yunfan Jiang, Ajay Mandlekar, Chaowei Xiao, Yuke Zhu,
  Linxi Fan, and Anima Anandkumar. 2023{\natexlab{a}}.
\newblock Voyager: An open-ended embodied agent with large language models.
\newblock \emph{arXiv preprint arXiv:2305.16291}.

\bibitem[{Wang et~al.(2025)Wang, Wang, Zhang, Yuan, Xu, tse Huang, Yuan, Guo,
  Chen, Wang, Xiao, and Zhou}]{wang2025cosercoordinatingllmbasedpersona}
Xintao Wang, Heng Wang, Yifei Zhang, Xinfeng Yuan, Rui Xu, Jen tse Huang, Siyu
  Yuan, Haoran Guo, Jiangjie Chen, Wei Wang, Yanghua Xiao, and Shuchang Zhou.
  2025.
\newblock \href {http://arxiv.org/abs/2502.09082} {Coser: Coordinating
  llm-based persona simulation of established roles}.

\bibitem[{Wang et~al.(2024)Wang, Xiao, Huang, Yuan, Xu, Guo, Tu, Fei, Leng,
  Wang et~al.}]{wang2024incharacter}
Xintao Wang, Yunze Xiao, Jen-tse Huang, Siyu Yuan, Rui Xu, Haoran Guo, Quan Tu,
  Yaying Fei, Ziang Leng, Wei Wang, et~al. 2024.
\newblock Incharacter: Evaluating personality fidelity in role-playing agents
  through psychological interviews.
\newblock In \emph{Proceedings of the 62nd Annual Meeting of the Association
  for Computational Linguistics (Volume 1: Long Papers)}, pages 1840--1873.

\bibitem[{Wang et~al.(2023{\natexlab{b}})Wang, Peng, Que, Liu, Zhou, Wu, Guo,
  Gan, Ni, Yang et~al.}]{wang2023rolellm}
Zekun~Moore Wang, Zhongyuan Peng, Haoran Que, Jiaheng Liu, Wangchunshu Zhou,
  Yuhan Wu, Hongcheng Guo, Ruitong Gan, Zehao Ni, Jian Yang, et~al.
  2023{\natexlab{b}}.
\newblock Rolellm: Benchmarking, eliciting, and enhancing role-playing
  abilities of large language models.
\newblock \emph{arXiv preprint arXiv:2310.00746}.

\bibitem[{Yang et~al.(2024)Yang, Yang, Zhang, Hui, Zheng, Yu, Li, Liu, Huang,
  Wei et~al.}]{yang2024qwen2}
An~Yang, Baosong Yang, Beichen Zhang, Binyuan Hui, Bo~Zheng, Bowen Yu,
  Chengyuan Li, Dayiheng Liu, Fei Huang, Haoran Wei, et~al. 2024.
\newblock Qwen2. 5 technical report.
\newblock \emph{arXiv preprint arXiv:2412.15115}.

\bibitem[{Ye(2024)}]{gusye1234_nanographrag}
Gu~Ye. 2024.
\newblock nano-graphrag.
\newblock \url{https://github.com/gusye1234/nano-graphrag}.
\newblock Accessed: 2025-06-22.

\bibitem[{Young et~al.(2024)Young, Chen, Li, Huang, Zhang, Zhang, Wang, Li,
  Zhu, Chen et~al.}]{young2024yi}
Alex Young, Bei Chen, Chao Li, Chengen Huang, Ge~Zhang, Guanwei Zhang, Guoyin
  Wang, Heng Li, Jiangcheng Zhu, Jianqun Chen, et~al. 2024.
\newblock Yi: Open foundation models by 01. ai.
\newblock \emph{arXiv preprint arXiv:2403.04652}.

\bibitem[{Yu et~al.(2022)Yu, Zhang, Xu, Lei, Guan, Zhang, Hou, Li, and
  Tang}]{yu2022xdai}
Jifan Yu, Xiaohan Zhang, Yifan Xu, Xuanyu Lei, Xinyu Guan, Jing Zhang, Lei Hou,
  Juanzi Li, and Jie Tang. 2022.
\newblock Xdai: A tuning-free framework for exploiting pre-trained language
  models in knowledge grounded dialogue generation.
\newblock In \emph{Proceedings of the 28th ACM SIGKDD Conference on Knowledge
  Discovery and Data Mining}, pages 4422--4432.

\bibitem[{Yu et~al.(2024)Yu, Yu, Wei, Zhang, and Qian}]{yu2024beyond}
Yeyong Yu, Runsheng Yu, Haojie Wei, Zhanqiu Zhang, and Quan Qian. 2024.
\newblock Beyond dialogue: A profile-dialogue alignment framework towards
  general role-playing language model.
\newblock \emph{arXiv preprint arXiv:2408.10903}.

\bibitem[{Zhang et~al.(2024)Zhang, Li, Tan, Yang, Zhu, Yang, Zhao, Ye, Li, and
  Hu}]{zhang2024cpsycoun}
Chenhao Zhang, Renhao Li, Minghuan Tan, Min Yang, Jingwei Zhu, Di~Yang, Jiahao
  Zhao, Guancheng Ye, Chengming Li, and Xiping Hu. 2024.
\newblock Cpsycoun: A report-based multi-turn dialogue reconstruction and
  evaluation framework for chinese psychological counseling.
\newblock \emph{arXiv preprint arXiv:2405.16433}.

\bibitem[{Zhong et~al.(2024)Zhong, Guo, Gao, Ye, and
  Wang}]{zhong2024memorybank}
Wanjun Zhong, Lianghong Guo, Qiqi Gao, He~Ye, and Yanlin Wang. 2024.
\newblock Memorybank: Enhancing large language models with long-term memory.
\newblock In \emph{Proceedings of the AAAI Conference on Artificial
  Intelligence}, volume~38, pages 19724--19731.

\bibitem[{Zhou et~al.(2023)Zhou, Chen, Wan, Wen, Song, Yu, Huang, Peng, Yang,
  Xiao et~al.}]{zhou2023characterglm}
Jinfeng Zhou, Zhuang Chen, Dazhen Wan, Bosi Wen, Yi~Song, Jifan Yu, Yongkang
  Huang, Libiao Peng, Jiaming Yang, Xiyao Xiao, et~al. 2023.
\newblock Characterglm: Customizing chinese conversational ai characters with
  large language models.
\newblock \emph{arXiv preprint arXiv:2311.16832}.

\bibitem[{Zhou* et~al.(2024)Zhou*, Zhu*, Mathur, Zhang, Qi, Yu, Morency, Bisk,
  Fried, Neubig, and Sap}]{zhou2024sotopia}
Xuhui Zhou*, Hao Zhu*, Leena Mathur, Ruohong Zhang, Zhengyang Qi, Haofei Yu,
  Louis-Philippe Morency, Yonatan Bisk, Daniel Fried, Graham Neubig, and
  Maarten Sap. 2024.
\newblock \href {https://openreview.net/forum?id=mM7VurbA4r} {Sotopia:
  Interactive evaluation for social intelligence in language agents}.

\bibitem[{Zimbardo and Ruch(1975)}]{zimbardo1975psychology}
Philip~G Zimbardo and Floyd~L Ruch. 1975.
\newblock Psychology and life.

\end{thebibliography}

    \clearpage

    \section*{Appendix}

    \appendix

    \section{Source Dataset}
    \vspace{-3pt}
    \label{sec: Source Dataset} 
    Our dataset was sourced from public English-language novel recommendations on platforms including The StoryGraph \citep{storygraph-browse}, Goodreads \citep{goodreads-fiction} and BookLikes \citep{booklikes-home}. An initial pool of 1,710 novels was compiled and subsequently filtered using a Large Language Model (LLM), resulting in a final selection of 539 novels that satisfy the following criteria: 

    \begin{itemize}
    \vspace{-3pt}
        \setlength{\itemsep}{1pt}
        \item Fluent, reader-friendly novel.
        \vspace{-3pt}
        \item Third-person narration.
        \vspace{-3pt}
        \item Clear formatting and well-structured layout.
        \vspace{-3pt}
        \item Balanced genre coverage to ensure diversity.
    \end{itemize}

    \section{Implementation Details}
    \vspace{-3pt}
    \subsection{Memory Module}
    \label{sub: Memory Module}

    We integrate Memory Module into the PsyMem framework to support contextual
    memory retrieval during role-playing. Specifically, we employed Nano-GraphRAG \citep{gusye1234_nanographrag}, an open-source lightweight library that implements a simplified version of GraphRAG \citep{edge2024local}. Since our test set consists of recently published novels, PsyMem relies on GraphRAG during inference to enrich the contextual content of the dialogue. We used a paragraph length of 8192 tokens for the memory text in local mode. Other configurations followed the default settings of Nano-GraphRAG. Specifically, in each dialogue turn, GraphRAG retrieves relevant entity information mentioned in the prior context to search for memory that aligns with the role the model is playing, and then combines this with character attributes to support inference.

    \subsection{Training Details}
    \label{sub:Training Details}

    We applied the PsyMem method to both Qwen2.5-7B-Instruct and LLaMA3.1-8B-Instruct models. During training, we incorporated data augmentation techniques to enhance the models’ ability to understand and generate content in role-playing tasks. Specifically, for each training instance, there was a 50\% probability of shuffling or masking the role attributes to expand the dataset. 

    \subsection{Hyperparameters}
    \label{sub:Hyperparameters}
    All models were trained on four A800 80G GPUs for 3 epochs with a global batch size of 128 and a sequence length of 8,196  (increased to 16,384 during the second training stage). LoRA was used with a per-device batch size of 2 and 16 gradient accumulation steps. We adopted a cosine learning rate schedule with a warmup ratio of 0.1, and set the learning rate to 1e-4.
    
    \section{Additional Evaluation}
    \vspace{-3pt}

    \begin{table}
        \centering
        \resizebox{\columnwidth}{!}{%
        \begin{tabular}{l|cccccc}
            \hline
            \multirow{2}{*}{\textbf{Model}} & \multicolumn{6}{c}{\textbf{Character Fidelity}} \\
            \cline{2-7}                     & \textbf{Per.}                                  & \textbf{Val.}  & \textbf{SL$^{*}$} & \textbf{BD$^{*}$} & \textbf{Mem.}  & {\textbf{Avg.}} \\
            \hline
            PsyMem-LLaMA (Rewrite)          & 80.14                                          & 81.47          & \textbf{78.62}    & 74.35             & 89.80          & 80.88           \\
            PsyMem-LLaMA (Random)           & 80.33                                          & 81.58          & 78.37             & \textbf{74.56}    & \textbf{90.20} & 81.00           \\
            PsyMem-LLaMA (Original)         & \textbf{80.47}                                 & \textbf{81.93} & 78.57             & 74.47             & \textbf{90.20} & \textbf{81.13}  \\
            \midrule                        
            PsyMem-Qwen(Rewrite)            & \textbf{80.70}                                 & 81.36          & 80.20             & 78.42             & 91.00          & 82.34           \\
            PsyMem-Qwen(Random)             & 80.45                                          & 81.63          & \textbf{81.02}    & 78.39             & \textbf{92.00} & \textbf{82.70}  \\
            PsyMem-Qwen(Original)           & 80.40                                          & \textbf{81.74} & 80.80             & \textbf{78.48}    & 91.80          & 82.64           \\
            \hline
        \end{tabular}%
        }
        \vspace{-10pt}
        \caption{Evaluation of the model’s generalization to input with freer
        phrasing (Rewrite), shuffled dimension order (Random), and prompts
        consistent with those used during training (Original).
        }
        \label{tab:generalization_study}
    \end{table}

    \subsection{More Flexible Inputs}
    To assess PsyMem's robustness to varied prompt styles, we tested three evaluation settings: the original structured prompt (`Original'), prompts with randomly shuffled character dimensions (`Random'), and character profiles rewritten into natural paragraphs by GPT-4o (`Rewrite'). Both PsyMem-LLaMA and PsyMem-Qwen were evaluated across five Character Fidelity dimensions.
    
    As shown in Table 5, the `Random' variant achieves comparable performance to the `Original' version, suggesting that the model is robust to prompt order perturbations (likely a result of training-time augmentation). The `Rewrite' variant results in a slight performance decrease (0.25\% for LLaMA, 0.30\% for Qwen), potentially due to minor information loss during rephrasing. Nonetheless, consistently high Memory scores across all variants demonstrate the robustness and generalization ability of PsyMem under varying input conditions.

    \subsection{Evaluation strategies}

    To verify the robustness of our evaluation methodology, we compared two scoring
    strategies: one strategy where each character's dialogues across different scenes were first concatenated and then assessed collectively, and another strategy where each character's dialogue, consisting of 15 turns per scene, was evaluated separately before being aggregated. We applied both strategies to four top-performing models, using
    three discriminative evaluators—GPT-4o \citep{hurst2024gpt}, OpenAI-o3 \citep{openai2025o3o4mini}, and Gemini-2.5-pro \citep{deepmind2025gemini2.5}. As shown
    in Table~\ref{tab:evaluation_strategy}, results across all models and
    evaluators exhibit strong consistency between the two strategies. Notably,
    for PsyMem-Qwen under GPT-4o evaluation, the average Character Fidelity score
    under concatenated dialogue input was 82.87\%, only 0.23\% points higher than
    the segment-wise evaluation score of 82.64\%, indicating that the choice of evaluation
    strategy does not significantly impact overall conclusions.

    \begin{table}[ht]
        \centering
        \resizebox{\columnwidth}{!}{
        \begin{tabular}{lccc}
            \hline
            \textbf{Model}       & \textbf{GPT-4o} & \textbf{o3}    & \textbf{Gemini-2.5-pro} \\
            \hline
            \textbf{GPT-4o}      & 80.86           & 79.62          & 76.92                   \\
            + Concatenated Eval. & 81.00           & 79.39          & 76.70                   \\
            \textbf{Deepseek-R1} & 82.47           & 78.40          & 75.87                   \\
            + Concatenated Eval. & 82.26           & 78.66          & 75.99                   \\
            \textbf{CoSER-70B}   & 82.28           & 80.23          & 79.16                   \\
            + Concatenated Eval. & 81.99           & 80.40          & 78.99                   \\
            \textbf{PsyMem-Qwen} & 82.64           & \textbf{80.74} & \textbf{79.70}                   \\
            + Concatenated Eval. & \textbf{82.87}  & 80.43          & {79.39}          \\
            \hline
        \end{tabular}
        }
        \vspace{-10pt}
        \caption{Average Character Fidelity scores (Per., Val., SL$^{*}$, BD$^{*}$,
        Mem.) of advanced role-playing models evaluated by different discriminative
        models and evaluation approaches. “Concatenated Eval.” refers to
        presenting all dialogues across scenarios simultaneously to the
        discriminative model for evaluation.
        }
        \label{tab:evaluation_strategy}
    \end{table}

    \begin{table}
        \centering
        \resizebox{\columnwidth}{!}{%
        \begin{tabular}{lcccc}
            \hline
            \textbf{Metrics}               & \textbf{Per.}   & \textbf{Val.}   & \textbf{SL$^{*}$} & \textbf{BD$^{*}$} \\
            \hline
            \textbf{Cosine} (H $\leftrightarrow$ L)      & 0.96            & 0.95            & 0.96              & 0.94              \\
            \textbf{\boldmath$r_{s}$} (H)  & 0.74            & 0.79            & 0.75              & 0.77              \\
            \textbf{\boldmath$\rho$} (H)   & 0.72            & 0.71            & 0.71              & 0.70              \\
            \textbf{Mean (Std)} (H)        & 0.59 $\pm$ 0.47 & 0.46 $\pm$ 0.42 & 0.42 $\pm$ 0.39   & 0.54 $\pm$ 0.41   \\
            \hline
            \textbf{\boldmath$\alpha$} (H) & 0.77            & 0.80            & 0.74              & 0.80              \\
            \hline
        \end{tabular}%
        }
        \vspace{-10pt}
        \caption{\textbf{Cosine} similarity, Pearson correlation coefficient (\textbf{\boldmath$\rho$}),
        Spearman’s correlation coefficient (\textbf{\boldmath$r_{s}$}), and
        \textbf{Mean (Std)} are used to assess the agreement and score distribution
        between human raters and GPT-4o across different evaluation dimensions.
        For reference, Krippendorff’s Alpha (\textbf{\boldmath$\alpha$}) is also
        reported as a measure of inter-rater reliability among human annotators.
        }
        \label{tab:Cosine similarity between human and GPT-4o scores}
    \end{table}

    \subsection{Human Evaluation}

    To validate our LLM-based evaluation framework, we conducted a large-scale
    human study involving 40 volunteers who assessed 25 dialogue scenarios (50 characters).
    The process was highly time-intensive, requiring over 10 hours per rater. For
    a robust analysis, we employed several complementary metrics. We used Krippendorff's
    Alpha ($\alpha$) to measure inter-rater reliability among human annotators,
    thereby establishing the reliability of our evaluation baseline. Concurrently,
    we assessed the agreement between GPT-4o's scores and the mean human scores
    using Cosine Similarity, Pearson's correlation coefficient ($\rho$), and
    Spearman's rank correlation coefficient ($r_{s}$).

    The results, summarized in Table~\ref{tab:Cosine similarity between human and GPT-4o scores}, validate our approach. We first confirmed a reliable human baseline, with Krippendorff's Alpha ($\alpha$) values of 0.74--0.80 indicating substantial inter-rater reliability \citep{krippendorff2018content, marzi2024k}. Against this baseline, GPT-4o's scores showed strong agreement with human ratings. Pearson's correlation ($\rho$) was consistently strong (0.70--0.72), and Spearman's rank correlation ($r_{s}$) was similarly high (0.74--0.79) \citep{fleiss1971measuring, hauke2011comparison}. These results collectively demonstrate that GPT-4o serves as a valid proxy for human evaluation in our task.

    \section{Evaluation-stage Input Ablation}
    \label{sub:Evaluation-stage Input Ablation}
    \vspace{-8pt}
    \begin{table}[ht]
        \centering
        \scriptsize
        \begin{tabular}{lcccc}
            \hline
            \textbf{Prompt Set} & \textbf{Per.} & \textbf{Val.} & \textbf{SL$^{*}$} & \textbf{BD$^{*}$} \\
            \hline
            \textbf{All}        & \textbf{80.40}         & \textbf{81.74}         & \textbf{80.80}             & \textbf{78.48}             \\
            \textbf{Per.}       & \underline{78.35}         & 73.33         & 74.72             & 73.67             \\
            \textbf{Val.}       & 74.74         & \underline{79.66}         & 74.35             & 74.39             \\
            \textbf{SL*}        & 68.12         & 67.86         & \underline{79.64}             & 69.35             \\
            \textbf{BD*}        & 69.38         & 68.77         & 70.14             & \underline{77.76}             \\
            \textbf{Base}       & 66.01         & 65.65         & 68.05             & 67.19             \\
            \hline
        \end{tabular}
        \vspace{-5pt}
        \caption{Performance of PsyMem-Qwen with different prompt combinations
        across four evaluation dimensions. “All” indicates inclusion of all prompts
        for the four dimensions; “Base” indicates exclusion of these prompts.
        Bold: Best; Underlined: Second best.
        }
        \vspace{-1pt}
        \label{tab:evaluation_qwen2.5}
    \end{table}
    
    We further investigated the role of character attributes during inference by
    varying the input format at evaluation time (as shown Table~\ref{tab:evaluation_qwen2.5}).
    Specifically, we compared three input conditions: (1) a minimal setting with
    no character dimensions, (2) inputs with only one active dimension, and (3)
    full inputs containing all four dimensions.

    The model’s performance deteriorated significantly when all dimensions were removed—for example, average scores dropped to as low as 66.01 on the Personality axis and 65.65 on Values. In contrast, even a single active dimension led to improved performance on the corresponding evaluation axis; using only the Personality prompt recovered the Personality score to 78.35, and similarly, Values alone yielded 79.66 on its dimension. 
    Additionally, we observed that providing Personality and Values information boosted performance on Social \& Leadership and Behavioral Decision dimensions (74.72 and 73.67 under Personality-only input), while the reverse was not true.
    This may suggest a hierarchical relationship among these attributes, where latent psychological attributes act as foundational traits that inform and guide performance in the more applied, explicit behavioral patterns.
\end{document}